\documentclass[letterpaper, 10 pt, conference]{ieeeconf} 

\usepackage{multirow}
\usepackage{cite}
\usepackage{amsmath,amssymb,amsfonts}
\usepackage{graphicx}
\usepackage{textcomp}
\usepackage{xcolor}
\usepackage{comment}
\usepackage{makecell} 
\usepackage{siunitx}
\usepackage{tikz}
\usetikzlibrary{
    arrows.meta,
    decorations.pathmorphing,
    shapes.geometric,
    calc,
    patterns
}
\usepackage{pgfplots}
\usepgfplotslibrary{groupplots}
\pgfplotsset{compat=1.18}
\usepackage{algorithm}
\usepackage{algpseudocode}
\usepackage{bm}
\usepackage{booktabs}
\usepackage{subcaption}
\usepackage[percent]{overpic}

\IEEEoverridecommandlockouts                          

\title{\LARGE \bf
Reactive Exploration of Unknown Environments \\ for Redundant Robots using Virtual Model Control}

\author{Alessio Canzolino$^{1}$, Omar Faris$^{2}$, Alessandro De Blasi$^{1}$, and Fulvio Forni$^{2}$
\thanks{This work was supported by the Engineering and Physical Sciences Research Council and AgriFoRwArdS CDT [EP/S023917/1].}
\thanks{$^{1}$Department of Electrical, Electronic, and Information Engineering ``Guglielmo Marconi'', University of Bologna, Italy}
\thanks{$^{2}$Department of Engineering, University of Cambridge, UK.}
}

\begin{document}

\maketitle
\thispagestyle{empty}
\pagestyle{empty}

\begin{abstract}

The exploration of confined, occluded, and partially known spaces poses significant challenges in robotic manipulation. The overall pose of the robotic arm must be carefully controlled to respect tight geometric constraints while avoiding newly discovered obstacles. We address this problem by proposing an active exploration approach for redundant robotic arms with an eye-in-hand camera configuration. Our approach navigates and acquires information in real-time based on a novel scoring method that directly selects a target voxel from the unexplored space using the robot's current state and expected information gain. To move the robot safely toward the target voxel, we utilize Virtual Model Control, which guarantees compliance and enables whole-body reactive obstacle avoidance without the need for path replanning. Simulated and real-robot experiments in both confined and open environments demonstrate the effectiveness of our approach, achieving over $90\%$ mapping coverage across all tested environments in under $120$ seconds without colliding with obstacles.

\end{abstract}

\section{Introduction}

The ability to explore unknown environments while acquiring information is a crucial goal of modern robotics and a fundamental enabler for deploying robots in everyday societal settings. These include search and rescue in unstructured environments \cite{sun2024survey}, inspection of hazardous settings \cite{wang2026dual}, and mapping of complex agricultural orchards \cite{zeng2022deep}. This classical problem has been extensively investigated for mobile robots \cite{lluvia2021active,Dobis2025,naazare2022online} but remains less explored in robotic manipulation. The use of robotic arms for active exploration presents unique challenges and risks, such as tight tolerances in confined spaces, occlusions, the non-trivial volume occupied by the arm's posture, and the complexity of physical interactions with multiple objects. These critical factors are particularly pronounced in robotic manipulation compared to mobile robotics, and lie at the core of nearly all manipulation tasks in industrial and domestic settings (such as reaching and grasping within shelves, boxes, drawers, etc.).

Active exploration with robotic arms requires movements that respect the robot's configuration and kinematics. Researchers typically address this by incorporating information about the arm posture and surrounding obstacles into motion planners that aim to reach a desired view \cite{Dobis2025, Butters2019, liu2024manipulability, renz2025next}. However, such methods are not optimized for confined, heavily occluded environments because planning latency prevents real-time, reactive motion around newly discovered obstacles. Furthermore, the choice between planned and reactive motion directly impacts sensing performance. To capture fine details in cluttered spaces, short-range depth sensors are preferred. These yield a small, local field of view that is ideal for reactive feedback control, but less effective for global trajectory optimization.

Our approach to active confined environment exploration with whole-body obstacle avoidance relies on 
two main components, both designed to take advantage of redundant robotic arms equipped with eye-in-hand camera configuration. First, we propose a novel scoring method that utilizes the arm posture to select the next target voxel within the unknown space. Existing methods \cite{Butters2019, liu2024manipulability, li2025compass} typically select views that maximize information gain over the unexplored region from a feasible set defined by the known space or its frontiers. In contrast, our scoring criterion evaluates voxels of the unknown space by ranking them using three metrics, relevance, frontality, and distance, which jointly account for the robot's kinematics and the expected information gain of each candidate voxel.
\begin{figure}[t]
\centering
\begin{tikzpicture}

    \node (main) {
        \includegraphics[width=0.9\linewidth]{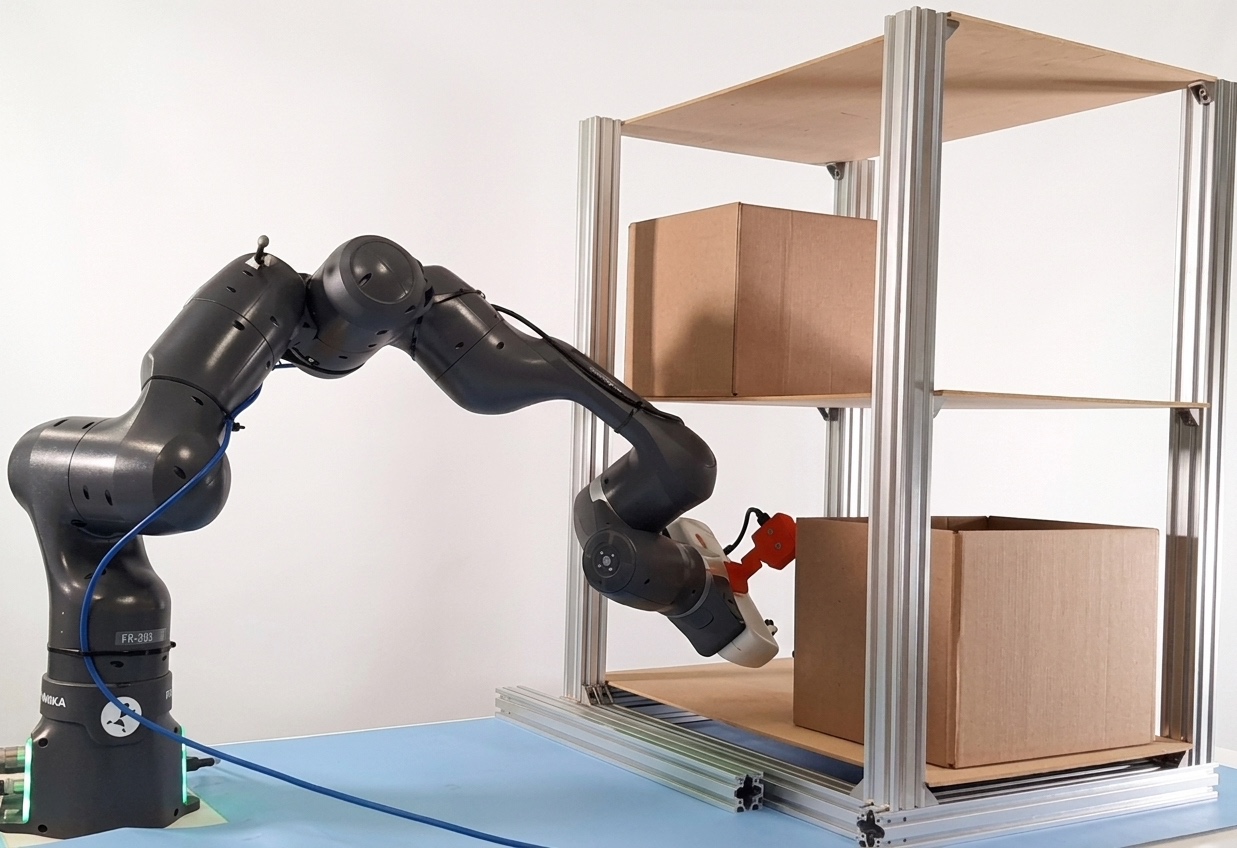}
    };

    \node[anchor=north east] at ([xshift=-1mm,yshift=-1mm]main.north east) {
    };

\end{tikzpicture}

\caption{The shelving unit for testing autonomous exploration in partially known environments.}
\label{fig:example}
\end{figure}
The second component is a control strategy based on Virtual Model Control \cite{Pratt2001, larby2025optimal}. The control action emulates a virtual mechanism attached to the robot, pulling it toward the target voxel in a reactive manner. This virtual mechanism serves as an interface between the robot and the environment, generating predictable, physics-driven motion without requiring path planning or explicit inverse kinematics computations. This enables whole-body obstacle avoidance and allows the robot to traverse narrow spaces by dynamically adjusting its posture.

We evaluate our approach across various unknown environments. The main test environment involves a three-tier shelving unit containing obstacles, shown in Fig. \ref{fig:example}, and a simulated benchmark of free-standing cubes with varying levels of clutter. These scenarios demonstrate active navigation through confined, partially known spaces while progressively acquiring environmental data. To highlight real-world applicability, we evaluate the system on additional two practical use cases: scanning three boxes containing internal objects and inspecting an artificial strawberry plantation with occluded fruits, which are tasks that require the robot to maneuver close to objects to capture fine structural details. The main contributions of this work are:
\begin{itemize}
    \item A novel criterion for selecting the next target voxel in the unexplored space.
    \item A new feedback controller, based on Virtual Model Control, that drives the robot towards the target with whole-body avoidance of newly discovered obstacles.
    \item Experimental validation on simulation and real robot across various confined and open environments.
\end{itemize}

\section{Background \& Related Works}

\subsection{Eye-in-hand Environment Exploration}

Autonomous exploration algorithms can generally be classified as frontier-based \cite{Yamauchi1997} or sampling-based \cite{bircher2016receding}. Frontier-based methods drive the robot towards the best unexplored region adjacent to the explored workspace (i.e., the ``frontier''). Sampling-based methods adapt conventional probabilistic path planners, such as RRT \cite{LaValle1998}, to exploration tasks, using them to sample views from the known free space. Modern exploration algorithms use these views to select a ``next best view'', defined as the view that maximizes the information gain about the unexplored regions \cite{Connolly1985}.

For robot arms with eye-in-hand camera configuration, researchers usually augment the next-best-view concept with the kinematic constraints of robotic arms to ensure feasible paths towards the chosen view. The algorithm proposed in \cite{paul2011autonomous} generates a collision-free path in the configuration space towards the next-best-view sensor pose. The algorithm was later improved in \cite{quin2013efficient} by conditioning the selection of the next-best-view to the feasible joint configurations adjacent to the current robot pose, resulting in faster exploration. In \cite{Butters2019}, the sensor pose was directly sampled in the Cartesian space rather than the configuration space, demonstrating efficient exploration in occluded and densely cluttered environments. The approach of \cite{Wang2019} 
selects the next view based on learnt motion direction and information gain quantification, and \cite{Dobis2025} relies on potential fields to generate candidate views to map the entire environment around the robot.

Other works focused on combining generation of next-best-views and motion planning. \cite{liu2024manipulability} 
derives the next-best-configuration from high manipulability metrics rather than solely maximizing the information gain. Similarly, \cite{li2025compass} plans the next-best-view in confined environments by balancing information gain, manipulability, and motion cost. \cite{renz2025next} derives the next-best-trajectory considering information gain metrics about the full trajectory rather than just the target view. \cite{zeng2022deep, ren2023robot, wang2026dual} rely on deep reinforcement learning to train next-best-view generation policies.

All these works focus on the efficient computation of next-best-views for poses reachable by the robotic arm along the frontier or within the known free space. While these approaches are well-suited for open environments, they appear less effective in confined spaces where the arm must traverse tight, cluttered areas. Our approach specifically targets this challenge. Instead of evaluating candidate poses that maximize information gain based on the existing map, we select the next target voxel using an aggregate score across all unknown voxels in the workspace. The robot is then driven directly toward this target voxel, relying on real-time reactive obstacle avoidance to navigate safely as new obstacles appear along the path.

\subsection{Virtual Model Control}

In Virtual Model Control \cite{Pratt2001,larby2025optimal}, the controller emulates virtual mechanisms attached to the robot that constrain and shape its motion. Virtual components, such as masses, springs, dampers, etc.  generate forces on the robot body according to physics. The concept was originally introduced for bipedal locomotion \cite{Pratt2001} and utilized later for quadrupedal locomotion \cite{chen2020virtual}. Virtual Model Control preserves the compliance of the robot motion and natively handles kinematic redundancy as it directly produces torque commands for the motors. These characteristics have made Virtual Model Control a successful methodology in collaborative and interactive tasks in unstructured environments, such as compliant reaching \cite{Zhang2024}, object handovers \cite{faris2026virtual}, vegetable cutting \cite{zhang2025periodic}, and collaborative drilling \cite{larby2025collaborative}.

In this work, we rely on Virtual Model Control for active environment exploration by designing virtual mechanisms that drive the robot towards the target voxel, eliminating the need for computationally demanding inverse kinematics. The robot is pulled toward the target voxel by virtual forces generated by contracting nonlinear springs and dampers attached between the camera and the target voxel. Concurrently, repulsive non-linear springs between newly discovered occupied voxels and the robot body govern whole-body collision avoidance during exploration, bypassing the need for motion planning algorithms. Motion planning and motion control are integrated. Interaction forces exerted by these virtual components render the target voxel an attractor and obstacles as repellers, guiding safe, reactive robot navigation.

\section{Environment Exploration Method}

We encode the information in the environment as a 3D voxel map in robot coordinate frame. The map is continuously updated based on the point cloud data from the depth camera attached to the robot end-effector. The exploration algorithm consists of three components: registration and update of voxel state, selection of the next target voxel, and exploration monitoring for deadlock management.

\subsection{Voxel State Registration and Update}

Each voxel can have one of three states: Unknown, Free, and Occupied. Each voxel stores an integer occupancy value $V(v) \in [V_{min}, V_{max}]$. For each camera frame, integer occupancy values for all voxels within the camera field-of-view are updated by incrementing or decrementing their values based on the camera point cloud, saturated by $V_{min}$ and $V_{max}$. The discrete state of a voxel $v$ is determined by two thresholds, $V_{\text{free}}$ and $V_{\text{occ}}$, such that:
\begin{equation}
    \text{State}(v) =
    \begin{cases}
        \text{Free},     & V_{\min} \le V(v) \le V_{\text{free}} \\
        \text{Unknown},  & V_{\text{free}} < V(v) < V_{\text{occ}} \\
        \text{Occupied}, & V_{\text{occ}} \le V(v) \le V_{\max}
    \end{cases}
\end{equation}
Due to noisy camera readings, we do not increment the integer occupancy of isolated voxels. This significantly reduces the effect of sensor noise, under the assumption that all obstacles are larger than the size of a single voxel.

\begin{figure}[t!]
  \centering
  \vspace{2.2mm}
  \resizebox{0.95\columnwidth}{!}{\begin{tikzpicture}[>={Stealth[length=2mm]}, font=\footnotesize,
    linkoutline/.style={line width=9pt, line cap=round, black},
    linkfill/.style={line width=7pt, line cap=round, teal!45},
    joint/.style={circle, draw=black, thick, fill=teal!20, minimum size=10pt, inner sep=0pt}]
\draw[gray!30, very thin, step=0.4] (0,0) grid (8,4.0);
\draw[thick] (0,0) rectangle (8,4.0);
\foreach \p in {(2.0,1.6),(2.4,1.6),(2.8,1.6),(2.0,2.0),(2.8,2.0),(2.0,2.4),(2.4,2.4),(1.6,2.0),(1.6,1.6),(2.4,2.8),
                (0.0,1.2),(0.0,1.6),(0.0,0.4),(0.4,1.6),(0.0,0.8),(0.4,1.2),
                (7.6,0.0),(7.2,0.0),(7.6,0.4),(7.2,0.4)}
  \fill[gray!45] \p rectangle ++(0.4,0.4);
\foreach \p in {(4.8,0.8),(5.2,0.8),(5.2,1.2),(4.8,1.2),(5.6,1.2),(5.6,1.6),(0.8,0.4),(0.8,0.8),(0.4,0.8),(4.4,0.8),(4.4,1.2),
                (6.4,3.2),(6.8,2.8),(6.8,3.2),(7.2,3.2),(0.4,2.8),(0.4,3.2),(0.4,3.6)}
  \fill[red!40] \p rectangle ++(0.4,0.4);
\foreach \p in {(3.2,1.2),(3.2,1.6)}
  \fill[red!40] \p rectangle ++(0.4,0.4);
\draw[very thick, orange!80!black, fill=orange!40] (2.4,2.0) rectangle ++(0.4,0.4);
\node[font=\footnotesize\itshape] at (2.6,2.2) {v};
\draw[-, very thick, green!60!black] (1.6,1.2) rectangle (3.6,3.2);
\node[green!60!black, align=left, anchor=south west] at (0.8,3.5)
    {$R(v)$: Relevance};
\draw[->, dashed, green!60!black] (2.3,3.6) -- (2.6,3.25);
\filldraw[teal!30, draw=black, thick] (3.7,-0.55) rectangle (4.3,-0.35);
\draw[teal!30, draw=black, thick] (3.85,-0.35) -- (3.85,-0.4) -- (4.15,-0.4) -- (4.15,-0.35) -- cycle;
\coordinate (J0) at (4,-0.35);     
\coordinate (J1) at (6.2,0.6);   
\coordinate (J2) at (6.6, 2.2);    
\coordinate (J3) at (5.8,3.2);    
\draw[linkoutline] (J0) -- (J1);
\draw[linkfill]    (J0) -- (J1);
\draw[linkoutline] (J1) -- (J2);
\draw[linkfill]    (J1) -- (J2);
\draw[linkoutline] (J2) -- (J3);
\draw[linkfill]    (J2) -- (J3);
\node[joint] at (J0) {};
\node[joint] at (J1) {};
\node[joint] at (J2) {};
\node[joint] at (J3) {};
\node[anchor=west] at (2.2,-0.4) {Robot base};
\draw[->, dashed, thick, blue!45!black] (4,-0.4) -- (2.75,2.05);
\node[blue!45!black, align=left, anchor=west] at (1.2,0.9)
    {$F(v)$: Frontality};
\begin{scope}[rotate around={20:(4.7,2.7)}]
  \fill[blue!15, draw=black, thick] (5.4,2.65) rectangle (5.9,2.95);
  \draw[thick, fill=blue!20]
    (5.4,2.72) -- (5.26,2.68) -- (5.26,2.92) -- (5.4,2.88);
\end{scope}
\node[anchor=west] at (4.5,3.5) {Camera};
\draw[->, dashed, thick, red!70!black] (5.2,3.0) -- (2.85,2.25);
\node[red!70!black, align=left, anchor=north west] at (3.9,2.75)
    {$D(v)$: Distance};
\fill[gray!45] (0.2,-1.2) rectangle ++(0.4, 0.4);
\node[anchor=west] at (0.6,-1.0) {unknown};
\fill[red!40] (2.4,-1.2) rectangle ++(0.4, 0.4);
\node[anchor=west] at (2.8,-1.0) {occupied};
\draw[very thick, orange!80!black, fill=orange!40] (4.6,-1.2) rectangle ++(0.3, 0.3);
\node[anchor=west] at (5.0,-1.0) {candidate target};
\end{tikzpicture}}
  \caption{2D visualization of the three scores used to select the next target voxel.
  $R(v)$ counts unknown voxels in the neighbourhood of $v$, $F(v)$ rewards voxels closer to the robot base, and $D(v)$ rewards voxels closer to the
  camera position.}
  \label{fig:target_scoring}
\end{figure}
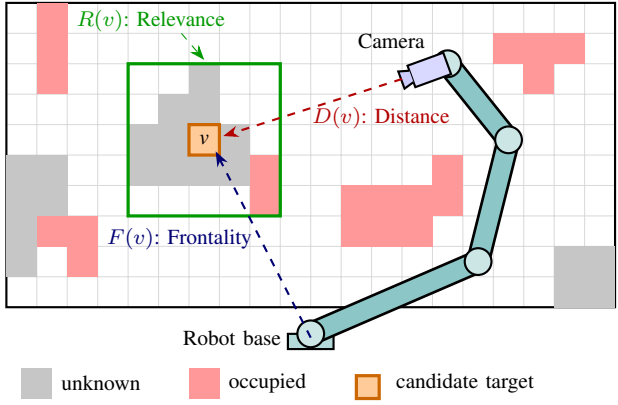

\subsection{Next Target Voxel Selection}

Three scores are associated to each unknown voxel $v$: Relevance $R(v)$, Frontality $F(v)$, and Distance $D(v)$. These are updated whenever a target voxel is in the field of view of the camera. The final score
\begin{equation}
  S(v) \;=\; w_R \, R(v) \;+\; w_F \, F(v) \;+\; w_D \, D(v)
  \label{eq:method_scoring}
\end{equation}
is used to select the next target voxel,
where $w_R, w_F$, and $w_D$ are user-defined (positive) weights. The voxel with the highest score is selected as the next target voxel.

The Relevance score, $R(v)$, is the number of unknown voxels within the local neighbourhood $N_{R(v)}$ of $v$ normalized by the total number of voxels in this neighbourhood. This score reflects the amount of information a voxel can provide if the robot reaches it. 
The Frontality score, $F(v)$, is proportional to the distance between the voxel $v$ and the robot base, $f_v$, normalized by the distance between the furthest edge of the workspace and robot base, $f_{max}$. Specifically, 
$F(v) = 1 - \frac{f_v}{f_{max}}$, such that the voxels closer to the robot base receive higher scores. Finally, the Distance score, $D(v)$, is defined as the distance $d_v$ between voxel $v$ and the camera position normalized to the maximum distance $d_{max}$ between the camera and the furthest voxel. Namely, $D(v) = 1 - \frac{d_v}{d_{max}}$. This ensures that voxels closer to the camera receive higher scores. Fig. \ref{fig:target_scoring} provides a graphical representation of the proposed score criterion.  

\subsection{Exploration Monitoring for Deadlock Management}

The robot control algorithm, detailed in the next section, guarantees controlled motion toward the unknown voxel alongside active obstacle avoidance. This is achieved without motion planning, by combining an attractive force pulling toward the target with repulsive forces exerted by obstacles. As a result, the robot may enter a deadlock state whenever the attractive force toward the target is balanced by the repulsive forces generated by obstacles. This can significantly slow down, or completely halt, the mapping task.

To ensure continuous exploration, we develop a deadlock monitoring module that flags a deadlock or near-deadlock state if either of two conditions are met: (i) the normalized discovery rate  $\gamma_d(t)$, defined as the rate at which new voxels are explored divided by the remaining unknown voxels, falls below a threshold $\lambda_{d}$ over a time window $\Delta t_{d}$, indicating unusual slow progress; or 
(ii) all joint velocities and applied torques fall below their respective thresholds, $\lambda_{vel}, \lambda_{torq}$, indicating a stagnant robot that cannot proceed to the target. 

If a deadlock is detected, the current target voxel is replaced by a new target. For the slow progress case (i), we select the second highest score target. This 
nudges the exploration towards voxels that are faster and easier to reach at the cost of a small score loss. For the stagnation case (ii),
we choose a random free voxel at a minimum distance from the original target, such that the robot is driven away from the latter.
Once the new random target is reached, the original target is reselected. If the arm reaches another deadlock on that target again (i.e., the same voxel produces two consecutive deadlocks), the target is declared unrecoverable.
In that case, we assume that the associated neighborhood area may cause further stagnation and, for efficiency, we establish a temporary ``ignored zone'' of fixed radius $r_{ign}$ around the unrecoverable voxel and exclude the area from future target selection.

\section{Virtual Mechanism for Robot Control}

\subsection{Overview}
The controller combines two virtual mechanisms: the \emph{camera nose driver} mechanism pulls the robot toward the target voxel, while the \emph{posture and obstacles regulator} mechanism generates repulsive forces between obstacles and the robot body to enable whole-body obstacle avoidance. Additionally, repulsive springs are activated near each joint limit to prevent the manipulator from exceeding its joint boundaries.

As is standard in virtual model control \cite{Pratt2001,larby2025optimal}, 
these virtual mechanisms are simulated and virtually connected to the
robot body at specific points $\mathbf{x}_i \in \mathbb{R}^3$, for $i \in \{1,\dots,n\}$, where they exert corresponding virtual
forces $\mathbf{F}_i$. These forces are then realized by the robot actuators (motors), via mapping into
generalized torques
\begin{equation}
    \boldsymbol{\tau}
    = \sum_{i=1}^{n}
    \mathbf{J}_{\mathbf{x}_i}(\mathbf{q})^T\,\mathbf{F}_i
    \label{eq:vmc_superposition}
\end{equation}
where $\mathbf{J}_{x_i}(\mathbf{q}) = \frac{\partial \mathbf{h}_{\mathbf{x}_i}(\mathbf{q})}{\partial \mathbf{q}}$ (Jacobian matrices). As long as the actuation stage accurately replicates the required torques, the robot behaves as if constrained and governed by its interconnection with the virtual mechanisms.

\subsection{Camera Nose Driver}

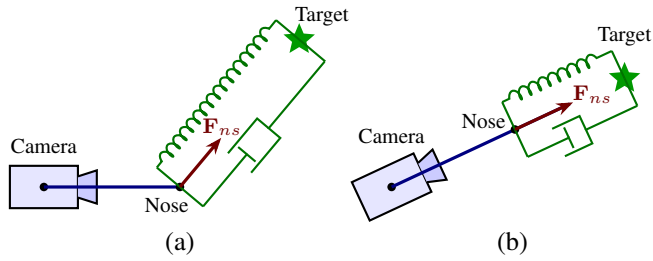
\begin{figure}[htbp]
  \centering
\begin{tikzpicture}[ >={Stealth[length=2.2mm]}, font=\footnotesize, spring/.style={thick, decorate, decoration={coil, segment length=4pt, amplitude=2.2pt, pre length=3pt, post length=3pt}} ]
\begin{scope} \node[font=\normalsize] at (1.8,-0.75) {(a)}; 
\draw[thick, fill=blue!10] (-0.45,-0.28) rectangle (0.45,0.28); \draw[thick, fill=blue!20] (0.45,-0.14) -- (0.7,-0.22) -- (0.7,0.22) -- (0.45,0.14); \fill (0,0) circle (1.6pt); \node[above=1pt] at (0,0.28) {Camera}; 
\draw[very thick, blue!50!black] (0,0) -- (1.8,0); \fill (1.8,0) circle (1.6pt); \node[below=0.5pt, xshift=-5.0pt] at (1.8,0) {Nose}; 
\node[
  star,
  star points=5,
  star point ratio=2.2,
  fill=green!60!black,
  inner sep=2pt
] (ta) at (30:3.9) {};

\node[above right=-5.0pt] at (ta.north) {Target};
\draw[thick, green!45!black]
  (0:1.8) -- ++(140:0.4);
\draw[thick, green!45!black]
  (0:1.8) -- ++(-40:0.4);
\draw[thick, green!45!black]
  (30:3.9) -- ++(140:0.4);
\draw[thick, green!45!black]
  (30:3.9) -- ++(-40:0.45);
\draw[spring, green!45!black]
  ($(0:1.8)+(140:0.4)$)
  --
  ($(30:3.9)+(140:0.4)$);
\draw[thick, green!45!black]
  ($(0:1.8)+(-40:0.4)+(50:0.9)$)
  -- ++(140:0.15);
\draw[thick, green!45!black]
  ($(0:1.8)+(-40:0.4)$)
  -- ++(50:0.9)
  -- ++(-40:0.15);
\draw[thick, green!45!black]
  ($(0:1.8)+(-40:0.4)+(50:0.75)+(-40:0.2)$)
  -- ++(50:0.6)
  -- ++(140:0.2)
  -- ++(50:1.15);
\draw[thick, green!45!black]
  ($(0:1.8)+(-40:0.4)+(50:0.75)+(140:0.2)$)
  -- ++(50:0.6)
  -- ++(-40:0.2);
\draw[->, very thick, red!45!black]
  (0:1.8) -- ++(50:0.85)
  node[above right, xshift=-11.0pt, yshift=-3.0pt, pos=1.0] {$\bm{\mathrm{F}}_{ns}$};
\end{scope} 
\begin{scope}[xshift=4.6cm] \node[font=\normalsize] at (1.6,-0.75) {(b)}; 
\begin{scope}[rotate=25] 
\draw[thick, fill=blue!10] (-0.45,-0.28) rectangle (0.45,0.28); 
\draw[thick, fill=blue!20] (0.45,-0.14) -- (0.7,-0.22) -- (0.7,0.22) -- (0.45,0.14); \end{scope} 
\fill (0,0) circle (1.6pt); 
\node[above=3pt] at (0,0.35) {Camera}; 
\draw[very thick, blue!50!black] (0,0) -- (25:2.6); 
\fill (25:1.8) circle (1.6pt); 
\node[above left, yshift=-3.0pt] at (25:1.8) {Nose}; 
\node[star, star points=5, star point ratio=2.2, fill=green!60!black, inner sep=2pt] (tb) at (25:3.4) {}; 
\node[above=2pt] at (tb.north) {Target}; 
\draw[thick, green!45!black] (25:1.8) -- ++(115:0.4);
\draw[thick, green!45!black] (25:1.8) -- ++(-65:0.4);
\draw[thick, green!45!black] (25:3.4) -- ++(115:0.4);
\draw[thick, green!45!black] (25:3.4) -- ++(-65:0.4);
\draw[spring, green!45!black] ($(25:1.8)+(115:0.4)$) -- ($(25:3.4)+(115:0.4)$); 
\draw[thick, green!45!black]
  ($(25:1.8)+(-65:0.4)+(25:0.6)$)
  -- ++(115:0.15);
\draw[thick, green!45!black]
  ($(25:1.8)+(-65:0.4)$) -- ++(25:0.6)
  -- ++(-65:0.15);
\draw[thick, green!45!black]
  ($(25:1.8)+(-65:0.4)+(25:0.45)+(-65:0.2)$) -- ++ (25:0.4) -- ++ (115:0.2) -- ++ (25:0.75);
\draw[thick, green!45!black]
  ($(25:1.8)+(-65:0.4)+(25:0.45)+(115:0.2)$) -- ++ (25:0.4) -- ++ (-65:0.2);
\draw[->, very thick, red!45!black] (25:1.8) -- ++(25:0.85) node[above right=-4.0pt] {$\bm{\mathrm{F}}_{ns}$}; 
\end{scope} 
\end{tikzpicture}
\caption{Camera nose mechanism: (a) configuration when a new target is selected (b) $\textbf{F}_{ns}$ pulls the nose in the direction of the target, orienting the camera accordingly.}
  \label{fig:nose_attractor}
\end{figure}

As illustrated in Fig. \ref{fig:nose_attractor},
the camera nose driver is modelled as a rigid ``nose'' protruding from the wrist camera through a virtual rigid link of length $l_n$, which connects to the target voxel via a virtual spring and damper. This enables simultaneous control of both camera's position and orientation, ensuring it points toward the voxel as it approaches. The nonlinear spring connecting the tip of the nose with the voxel generates saturated force of the form
\begin{equation}
    \label{eq:force_profile}
    \textbf{F}_{ns} = \bar{F}_{{ns}} \tanh\!\left(\frac{k_{ns}|\textbf{x}|}{\bar{F}_{{ns}}}\right)\frac{\textbf{x}}{|\textbf{x}|}
\end{equation}
where $\bar{F}_{{ns}}>0$ is the maximum force exerted by the spring and $k_{ns}>0$ is its stiffness (we take the limit $\textbf{F}_{ns}=0$ if $\textbf{x}=0$). The linear damper takes the usual form
\begin{equation}
\textbf{F}_{nd} = c\dot{\textbf{x}}.
\end{equation}
where $c>0$ is the damping coefficient.
\begin{figure}[t!]
    \centering
    \vspace{2.2mm}
    \includegraphics[width=0.8\linewidth]{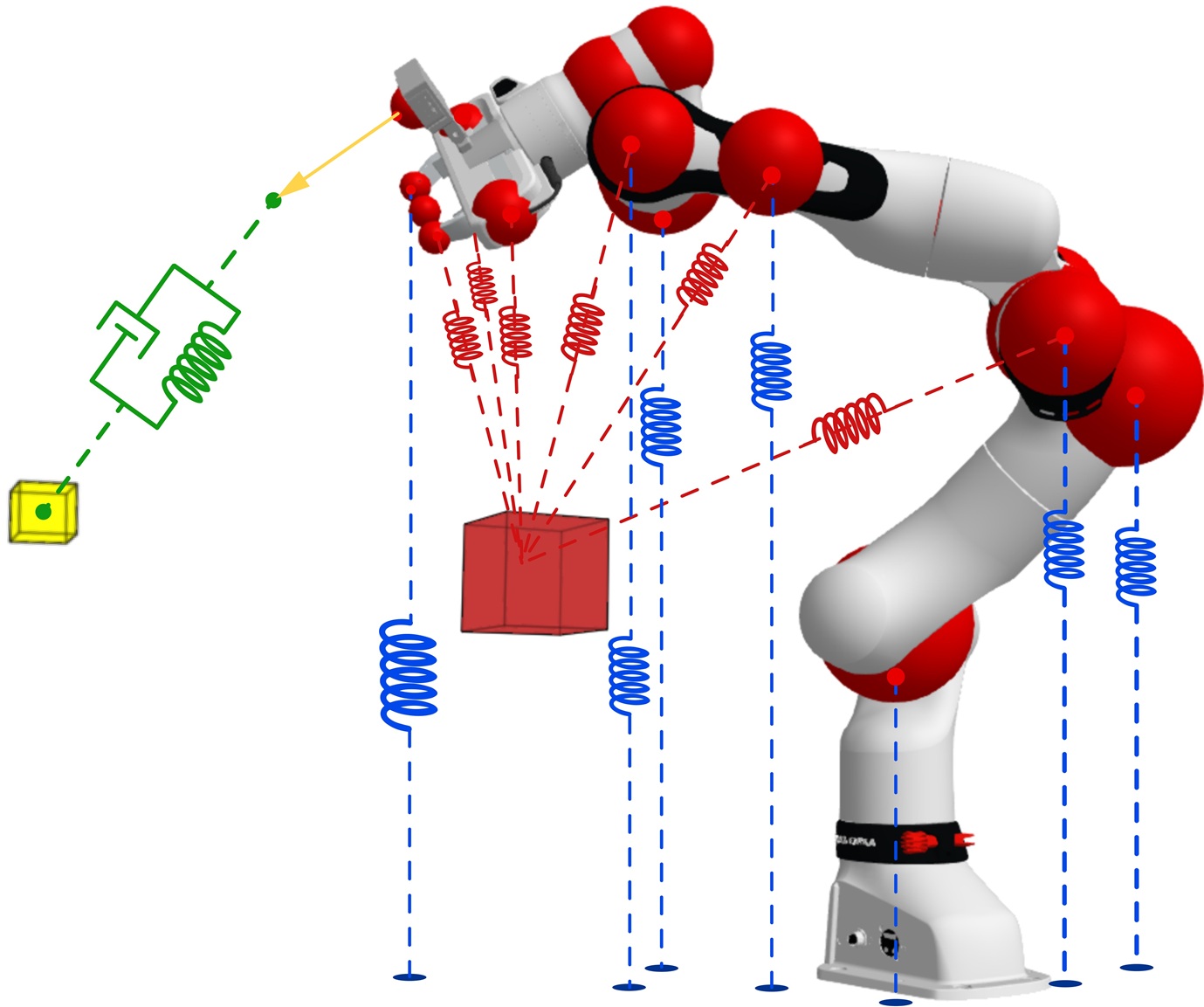}
    \caption{The green spring and damper pull the camera nose (orange arrow) towards the target (yellow box). Repulsive springs from obstacles (red) and base plane (blue) regulate the robot posture, enabling whole-body obstacle avoidance. 
    }
    \label{fig:vmc_mechanism}
\end{figure}

\subsection{Posture and Obstacles Regulator}

Fig. \ref{fig:vmc_mechanism} illustrates the working principle of the posture and obstacles regulator mechanism. 
As the robot explores the environment, virtual repulsive nonlinear springs (red) are continuously added between occupied voxels and pre-defined collision avoidance points distributed along the robot body. Additional
repulsive nonlinear springs (blue) connect the robot to the base plane to avoid collisions. 

The force profile of the $i$-th spring is derived from the Gaussian potential,
\begin{equation}
   \textbf{F}_{r,i} = -|\textbf{x}_{r,i}|k_{r,i}\exp\left(\frac{-|\textbf{x}_{r,i}|^2}{2\sigma_{r,i}^2}\right)
    \label{eq:repulsive}
\end{equation}
\noindent where $|\textbf{x}_{r,i}|$ is the extension of the $i$-th spring, $k_{r,i}>0$ define the intensity of the force given a certain spring extension, while $\sigma_{r,i}>0$ practically determines the range of the repulsive force.

\begin{figure*}[htbp]
    \centering
    \vspace{2.2mm}
    \begin{subfigure}[t]{0.34\textwidth}
        \centering
        \includegraphics[width=\linewidth]{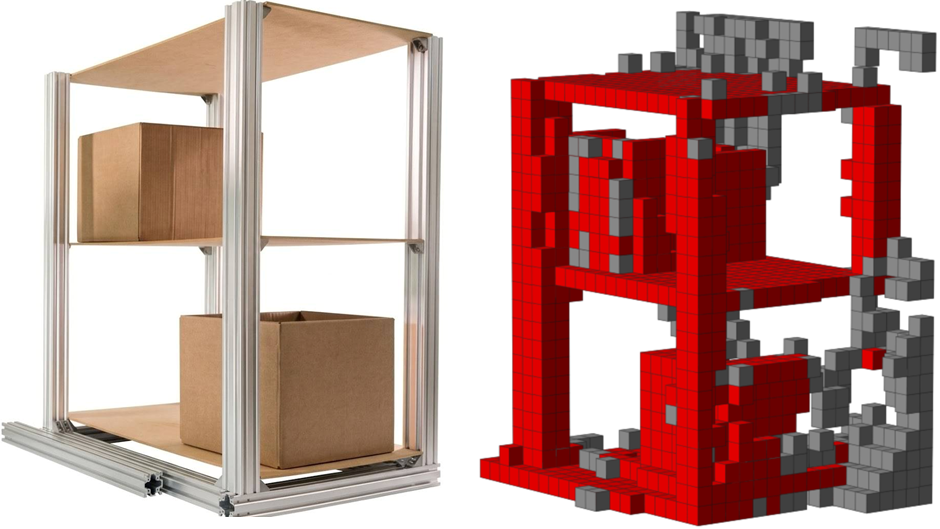}
        \caption{}
        \label{fig:shelves_realvsmapped}
    \end{subfigure}
    \hspace{0.002\textwidth}
    \begin{subfigure}[t]{0.31\textwidth}
        \centering
        \includegraphics[width=\linewidth]{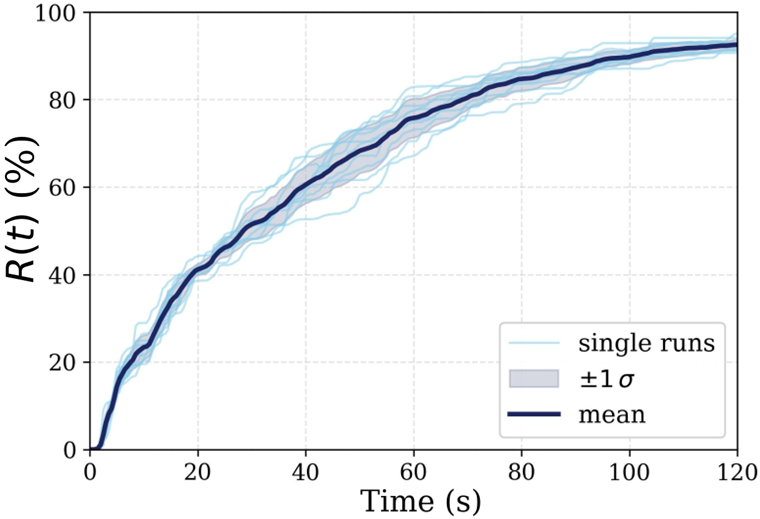}
        \caption{}
        \label{fig:shelves_rt}
    \end{subfigure}
    \hspace{0.002\textwidth}
    \begin{subfigure}[t]{0.32\textwidth}
        \centering
        \includegraphics[width=\linewidth]{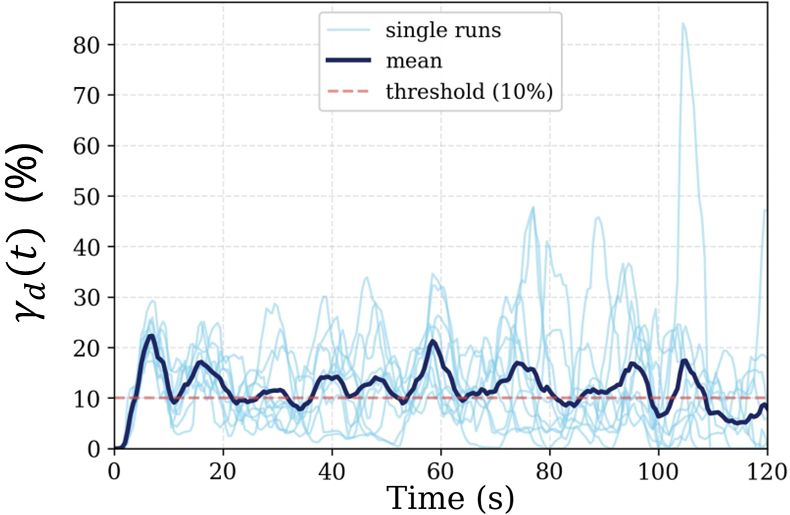}
        \caption{}
        \label{fig:shelves_drm}
    \end{subfigure}

    \caption{(a) Real vs mapped shelving unit from one experimental run. Red voxels have occupied status ($V_{occ}$) and gray voxels are unexplored ($V_{unk}$). (b) Exploration curve $R(t)$ for all experimental runs. (c) Curves of the normalized discovery rate $\gamma_d(t)$ accumulated over a rolling time window of $5$~s for all experimental runs.}
    \label{fig:shelves_results}
\end{figure*}

\section{Exploration of Confined Environments}

\subsection{Setting and Evaluation Metrics}
We evaluate our exploration algorithm in confined environments. The primary experimental environment is a three-tier shelving unit containing internal boxes. Navigating this environment requires complex maneuvers for the robot arm to reach between shelves and thoroughly scan the workspace. Additional simulations provide further testing for cluttered and occluded environments.

All experiments were conducted using a Franka FR3 robotic arm with an eye-in-hand Intel RealSense D405 depth camera. The specifications of the camera were also replicated in simulation tests. The controller was implemented using the VMRobotControl.jl package \cite{VMRobotControl}. ROS 2 was used as the middleware to communicate among all the hardware components, whereas all software programs run on a CPU. 

\begin{table}[b!]
  \centering
  \begin{tabular}{lc|lc}
    \toprule
    Parameter & Value & Parameter & Value \\
    \midrule
    $w_R,w_F,w_D$ & $2,1,4$ &
     $l_n$ & $25$~cm  \\
    $V_{min}, V_{max} $ & $ 0, 350  $ &  $k_{ns} $& $300$~N/m  \\
    $ V_{free}, V_{unk}, V_{occ}$ & $20, 55, 90$ & $k_{r,i}$ & $330$~N/m  \\
    $\lambda_{vel}, \lambda_{torq}$ & $0.15$~m/s, $0.02$~Nm & $\bar{F}_{ns}$ & $18$~N \\  
     $r_{ign}$ & $5$& $c$ & $8.5$~Ns/m \\
    $N_{R(v)}$ & $3$ & $\sigma_{r,i}$ & $5$~cm \\
    $\lambda_d$, $\Delta t_d$& $10\%$, $5$~s & &  \\
    \bottomrule
  \end{tabular}
  \caption{Parameters used in experiments and simulations.}
  \label{tab:params}
\end{table}

Both the simulation and physical experiments used the same parameters, provided in Table \ref{tab:params}. Mechanical parameters were selected empirically. 
Scoring criterion parameters and weights were chosen to prioritize exploration of voxels closer to the robot. This choice was found to be suitable for cluttered environments as it minimizes the risk of collisions with unknown spaces. Although a systematic optimization of these parameters would improve performance, it is beyond the scope of this paper. We refer interested readers to \cite{larby2025optimal} for a structured approach to parameter optimization in Virtual Model Control.

We evaluate the performance of our algorithm using metrics commonly adopted in previous works \cite{Butters2019, liu2024manipulability}. We report:
(i) the percentage of explored workspace
  $R_\infty$ at the end of each exploration, which is terminated if the exploration is complete or after $120$ seconds; 
(ii) the exploration time $T_e$, defined as the time required to explore $95\%$ of $R_\infty$;
(iii) the total path length $L$ traveled by the camera during the exploration;
(iv) the number of times a new target has been selected, $N_{ts}$, which we further specialize into targets reached $N_{ts:TR}$, discovery rate deadlocks $N_{ts:\gamma}$, and stagnation deadlock $N_{ts:\lambda}$;
(v) the proportion of time the robot has been stagnant, $p_{stag}$, recorded whenever all joints velocities are below $\lambda_{vel}$ or all applied torques are below $\lambda_{torq}$; 
and (vi) the number of observed deadlocks $N_{dl}$. We also provide plots of
the percentage of explored workspace
 at time $t$, $R(t)$.

\begin{table}[b!]
  \centering
  \begin{tabular}{lc|lc}
    \toprule
    Metric & Mean $\pm$ SD & Metric & Mean $\pm$ SD \\
    \midrule
    $R_\infty$ (\%) & $92.5 \pm 1.3$ & $T_e$ (s) & $91.4 \pm 7.3$ \\
    $L$ (m)          & $13.4 \pm 1.0$ & $N_{\mathrm{ts}}$ & $47.4 \pm 4.8$ \\
    $p_{\mathrm{stag}}$ (\%) & $40.9 \pm 6.1$ & $N_{ts:TR}$ & $28.8 \pm 3.4$ \\
    $N_{\mathrm{dl}}$ & $29.8 \pm 2.1$ & $ N_{ts:\gamma}\,,\,N_{ts:\lambda}$ & $12.0 \pm 1.5$\,,\, $6.6 \pm 4.6$  \\
    \bottomrule
  \end{tabular}
  \caption{Experimental results for the exploration of three-tier shelving unit environment.}
  \label{tab:exp_shelf_kpi}
\end{table}

\subsection{Three-tier Shelving Unit Experiment}

We conducted 10 experimental runs on the three-tier shelving unit environment of Fig. \ref{fig:example}. The voxel resolution was set at $3.5$ cm,
 generating 12075 voxels within the shelving unit volume $(x,y,z) \in [-0.25\mathrm{m},\,0.55\mathrm{m}]\times[0.40\mathrm{m},\,1.11\mathrm{m}]\times[0.03\mathrm{m},\,0.90\mathrm{m}] $, relative to the robot base coordinate frame
 (smaller voxel resolutions were constrained by the available computational power).

\begin{figure*}[t!]
    \centering
    \vspace{2.2mm}
    \begin{subfigure}[b]{0.22\textwidth}
        \centering        
        \includegraphics[width=1\linewidth]{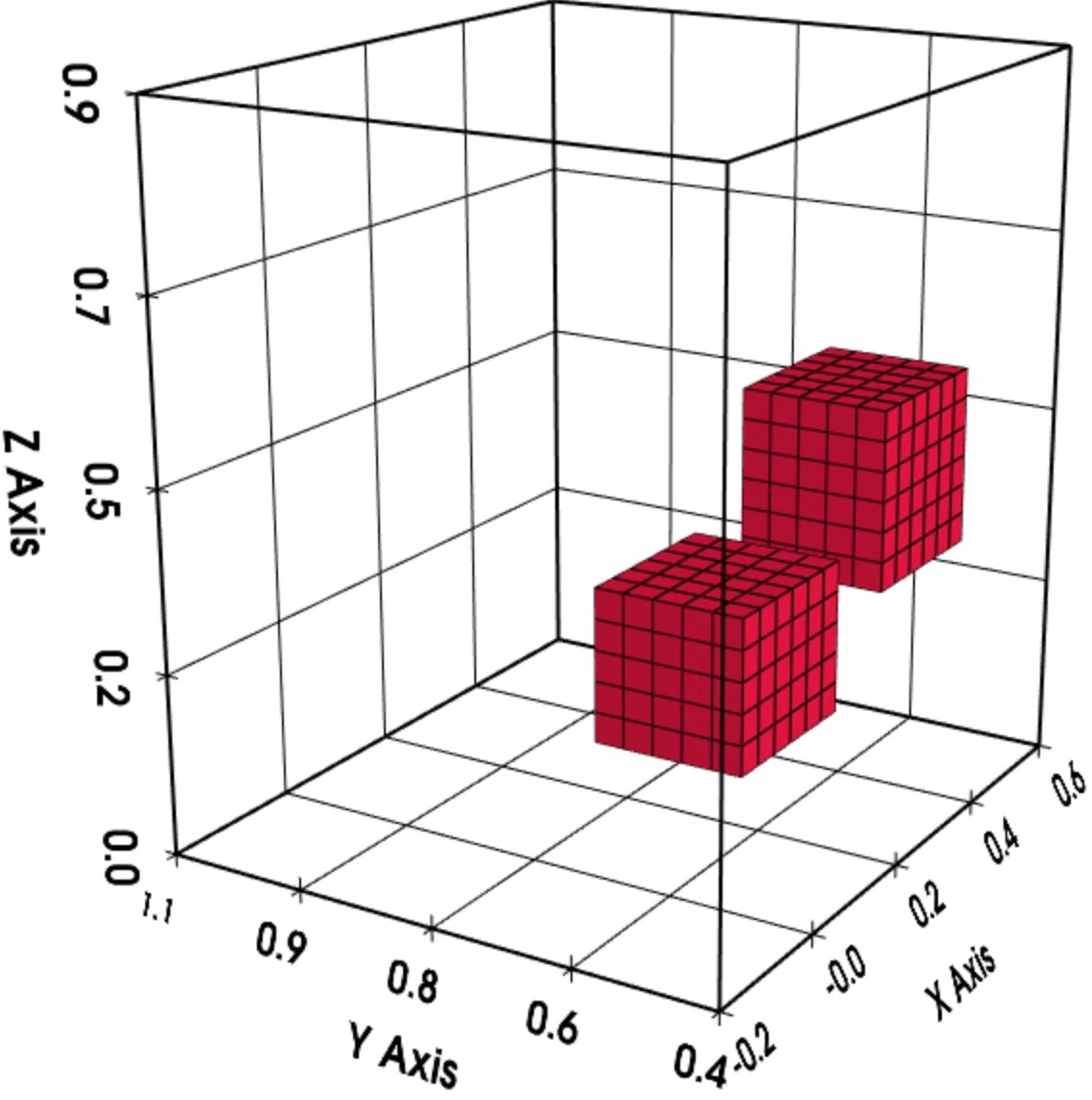}
        \caption{}
        \label{fig:scene_a}
    \end{subfigure}
    \hspace{0.01\textwidth}
    \begin{subfigure}[b]{0.23\textwidth}
        \centering
        \includegraphics[width=\textwidth]{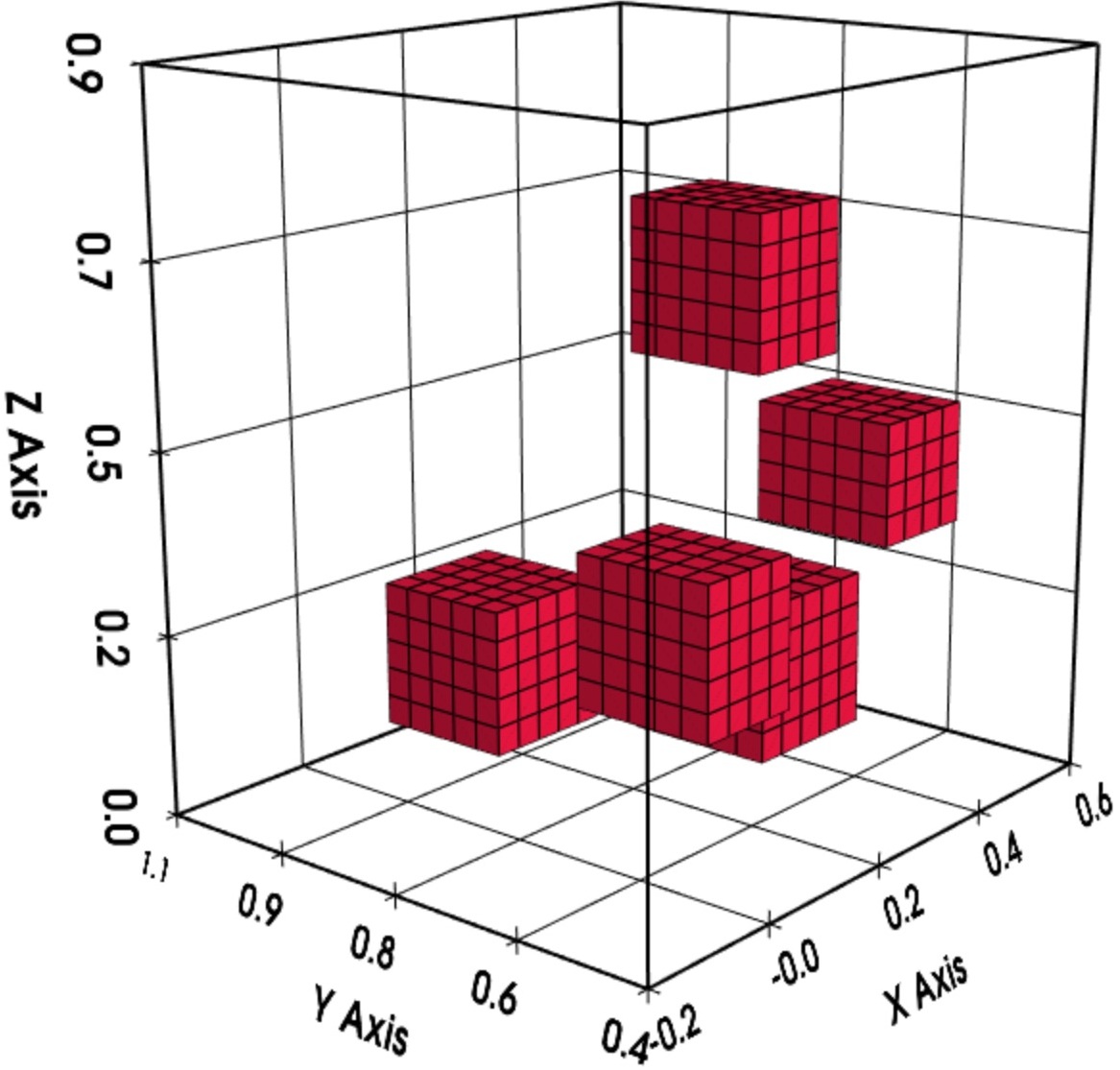}
        \caption{}
        \label{fig:scene_b}
    \end{subfigure}
    \hspace{0.01\textwidth}
    \begin{subfigure}[b]{0.23\textwidth}
        \centering
        \includegraphics[width=\textwidth]{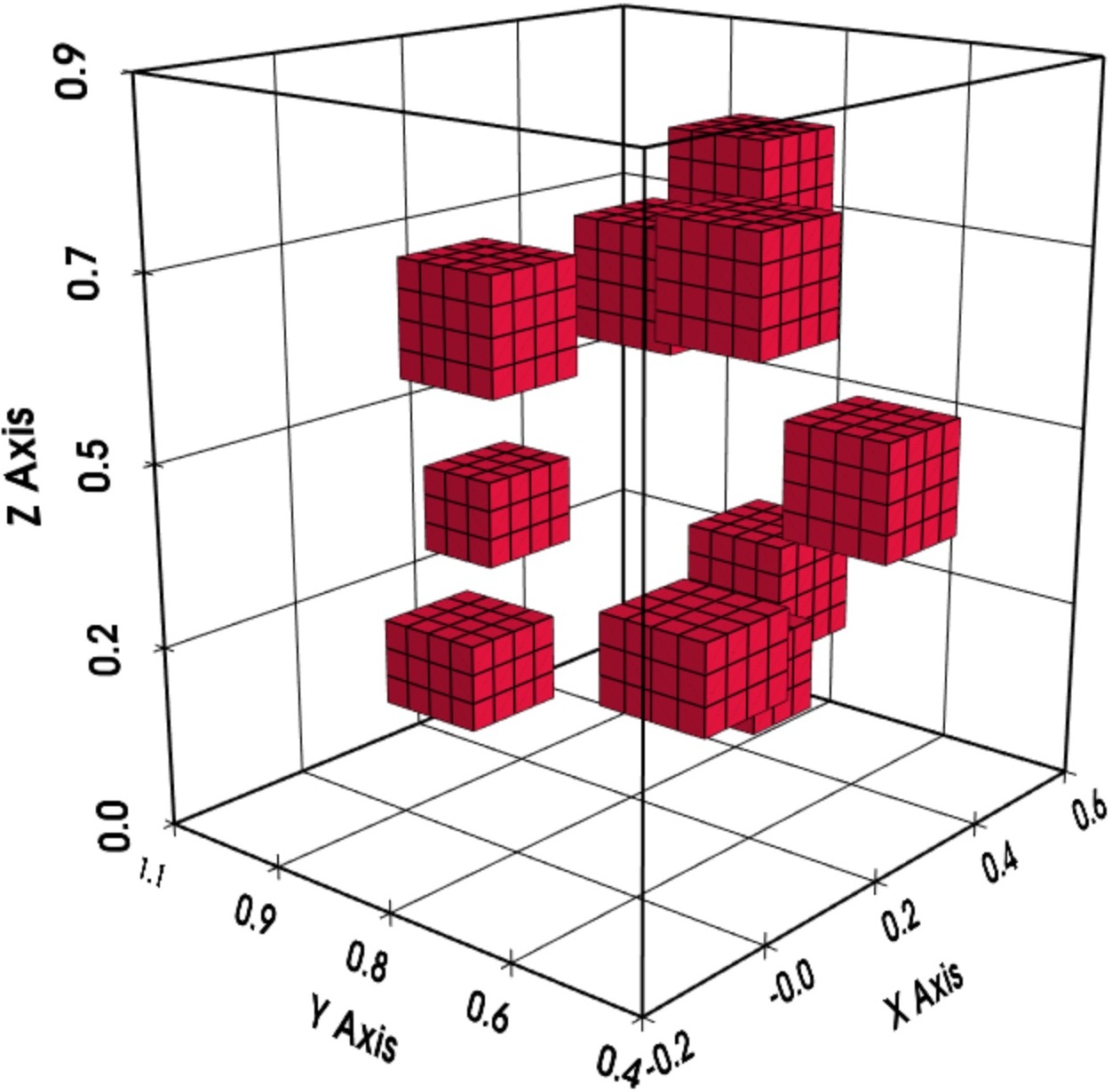}
        \caption{}
        \label{fig:scene_c}
    \end{subfigure}
    \hspace{0.01\textwidth}
    \begin{subfigure}[b]{0.22\textwidth}
        \centering
        \includegraphics[width=\textwidth]{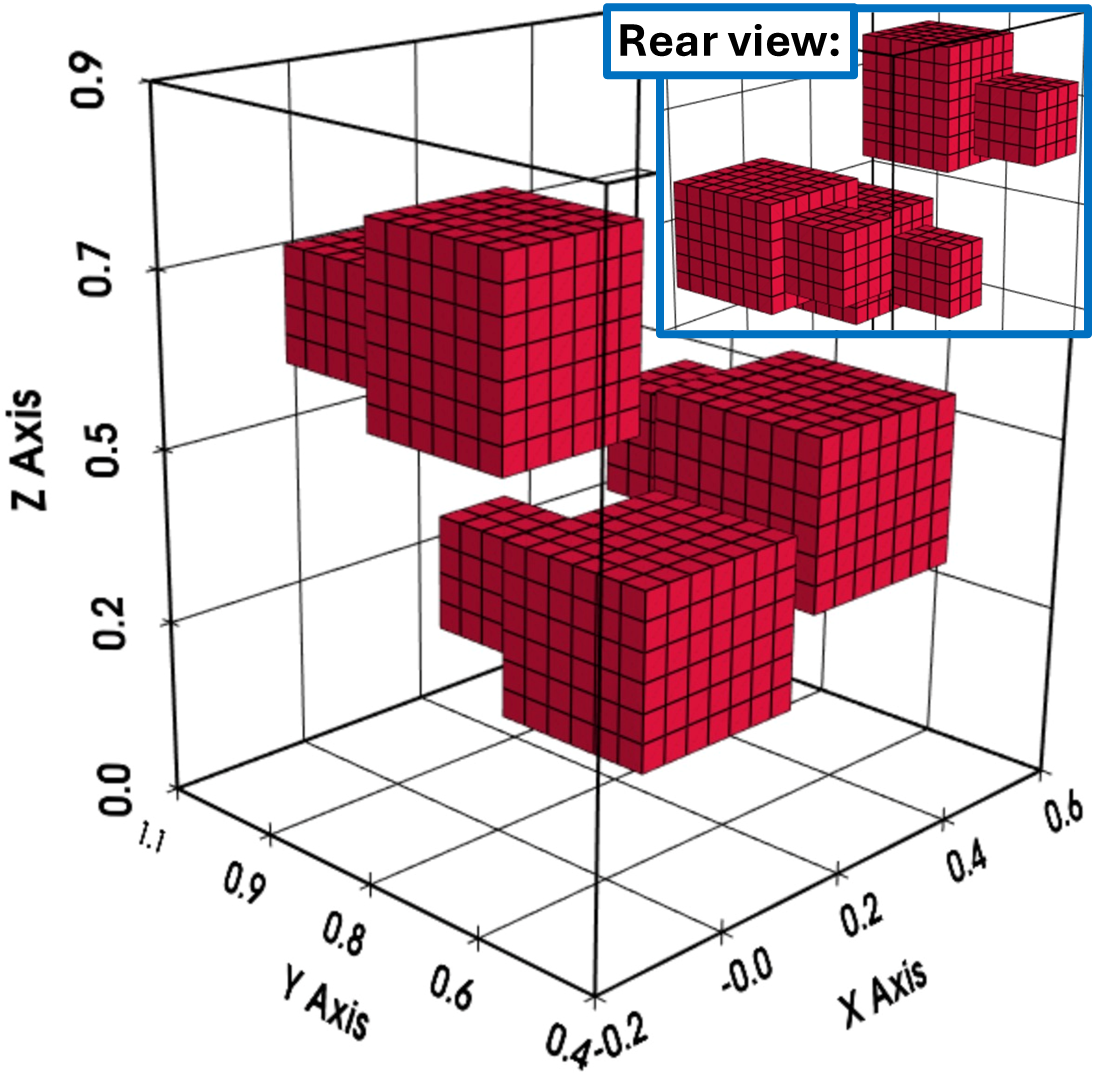}
        \caption{}
        \label{fig:scene_d}
    \end{subfigure}
    
    \caption{Simulation tests with increased levels of clutter. Scenarios A-C shows increased number of smaller obstacles. Scenario D highlights challenging occlusions, as shown in the rear view.}
    \label{fig:abc}
\end{figure*}

\begin{figure*}[t!]
    \centering

    \begin{subfigure}[t]{0.32\textwidth}
        \centering
        \includegraphics[width=\textwidth]{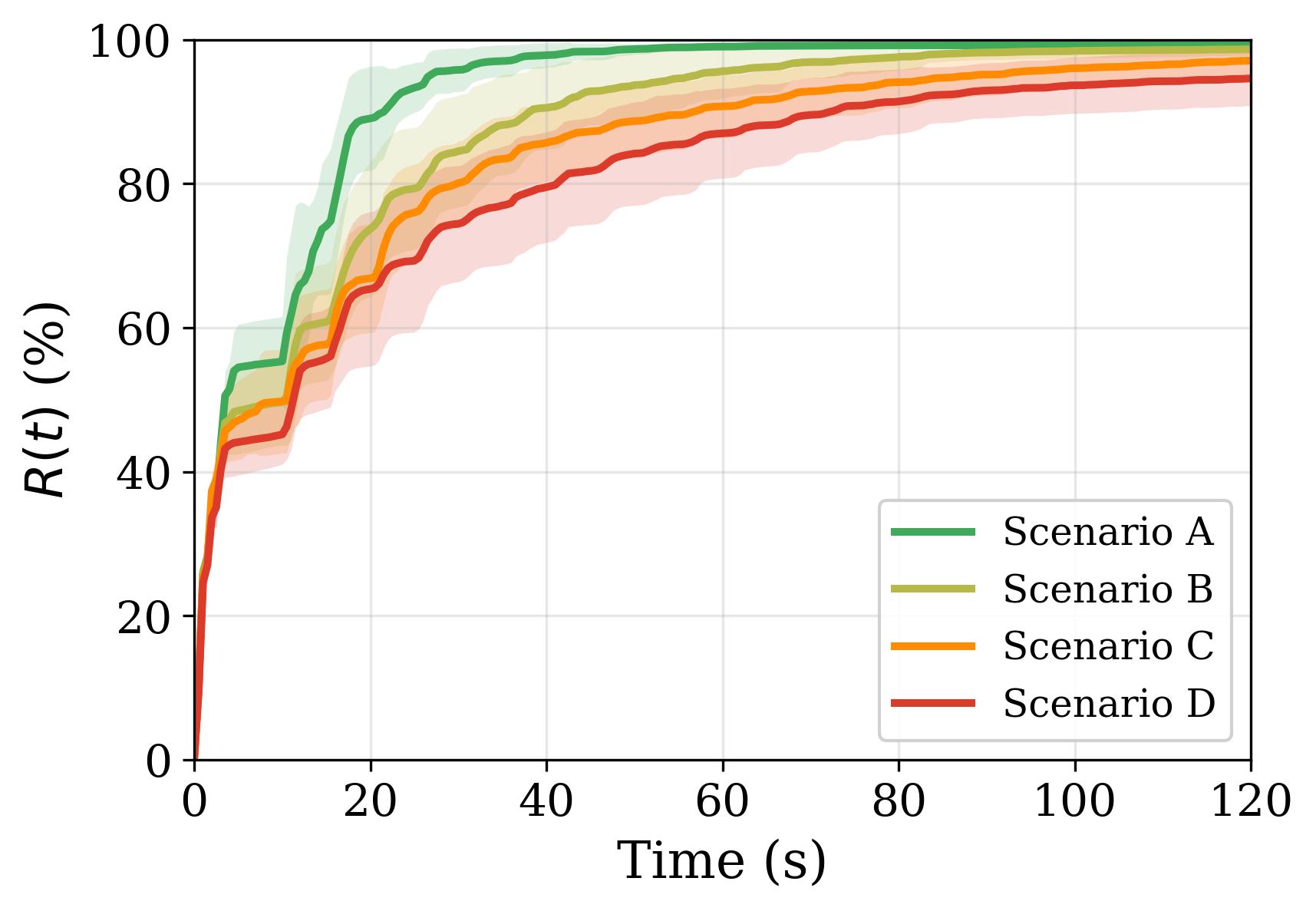}
        \caption{}
        \label{fig:phase2_discovery}
    \end{subfigure}
    \hfill
    \begin{subfigure}[t]{0.42\textwidth}
        \centering
        \includegraphics[width=\textwidth]{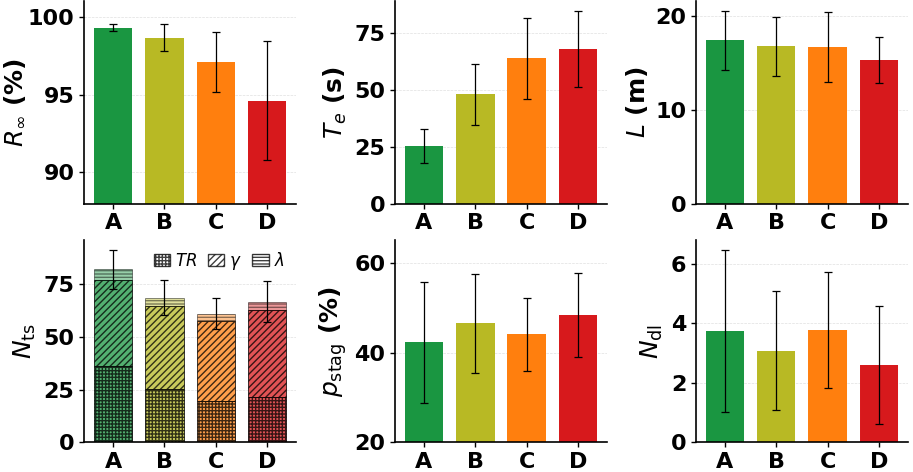}
        \caption{}
        \label{fig:simulation_results}
    \end{subfigure}
    \hfill
    \begin{subfigure}[t]{0.22\textwidth}
        \centering
        \includegraphics[width=\textwidth]{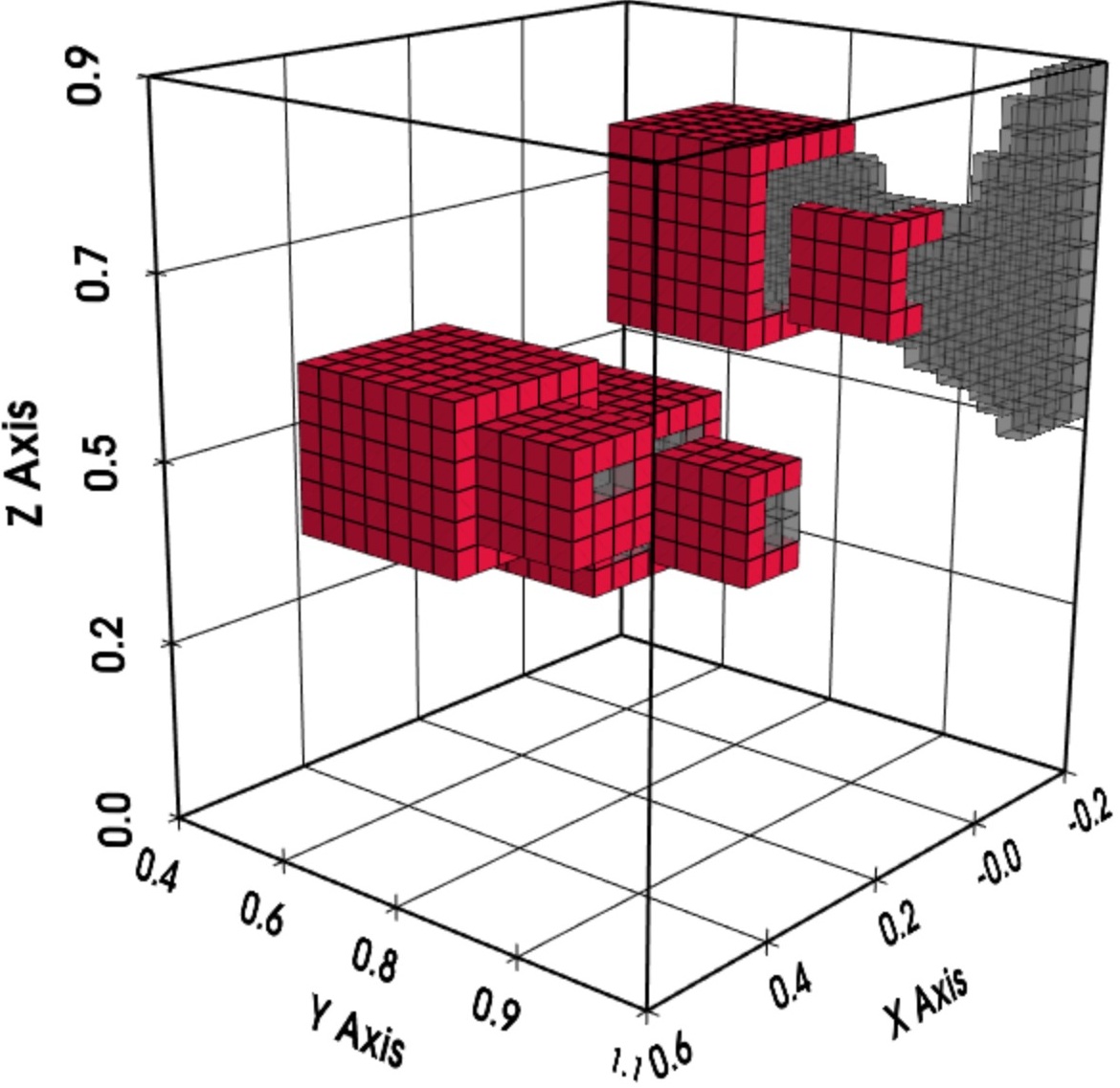}
        \caption{}
        \label{fig:rear_D}
    \end{subfigure}

    \caption{Simulation results for the Scenarios A, B, C, and D, with $R(t)$ shown in (a), the remaining metrics shown in (b), and the rear view of a mapped Scenario D environment shown in (c).}
    \label{fig:combined_results}
\end{figure*}

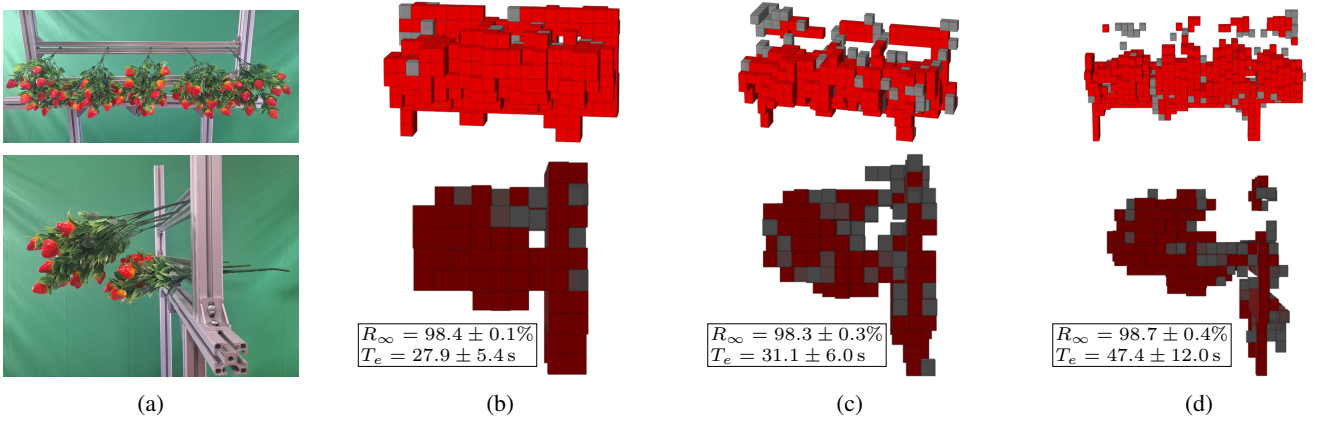
\begin{figure*}[t]
  \centering
  \begin{subfigure}[t]{0.22\textwidth}
    \centering
    \includegraphics[width=\textwidth]{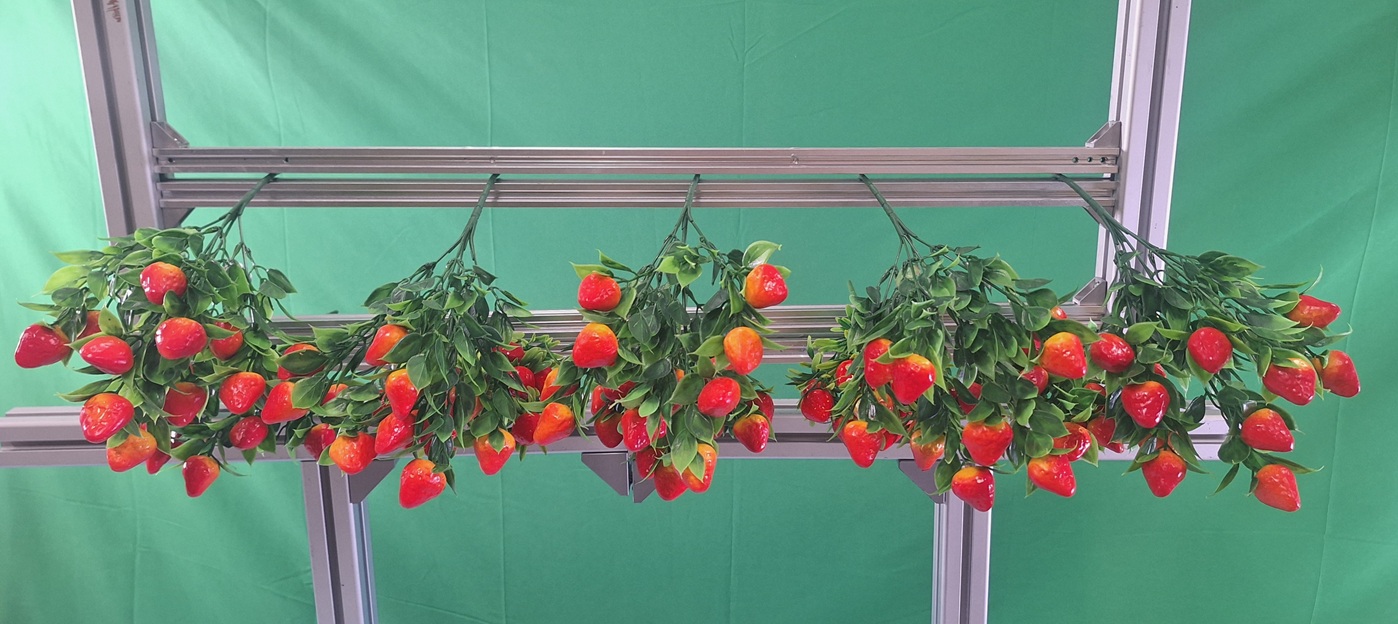}
    \label{fig:g32_a}
  \end{subfigure}
  \hfill
  \vspace{-0.8em}
  \begin{subfigure}[t]{0.22\textwidth}
    \centering
    \includegraphics[width=0.8\textwidth]{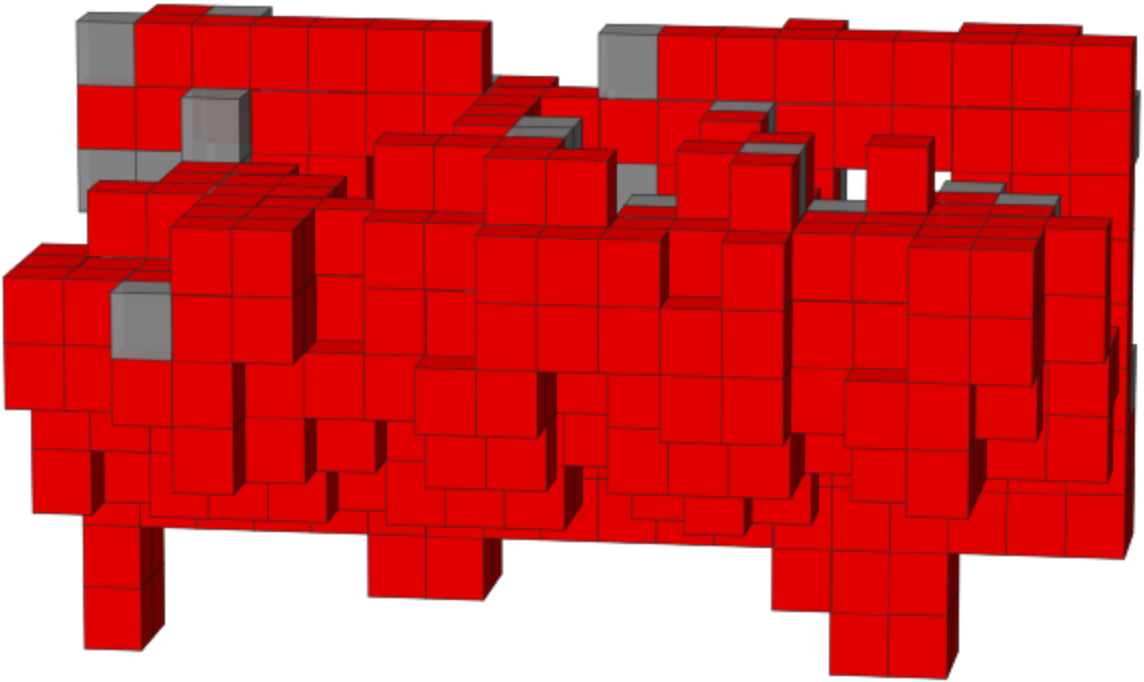}
    \label{fig:g32_b}
  \end{subfigure}
  \hfill
  \begin{subfigure}[t]{0.22\textwidth}
    \centering
    \includegraphics[width=0.75\textwidth]{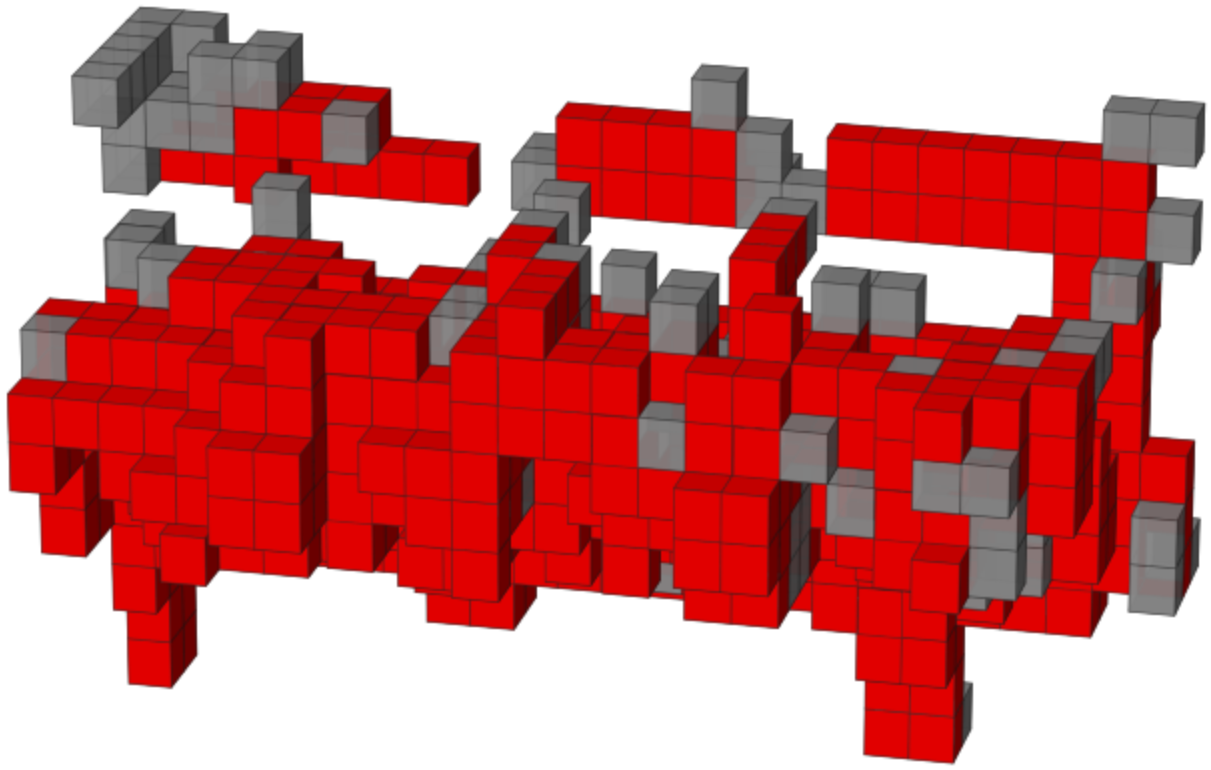}
    \label{fig:g32_c}
  \end{subfigure}
  \hfill
  \begin{subfigure}[t]{0.22\textwidth}
    \centering
    \includegraphics[width=0.75\textwidth]{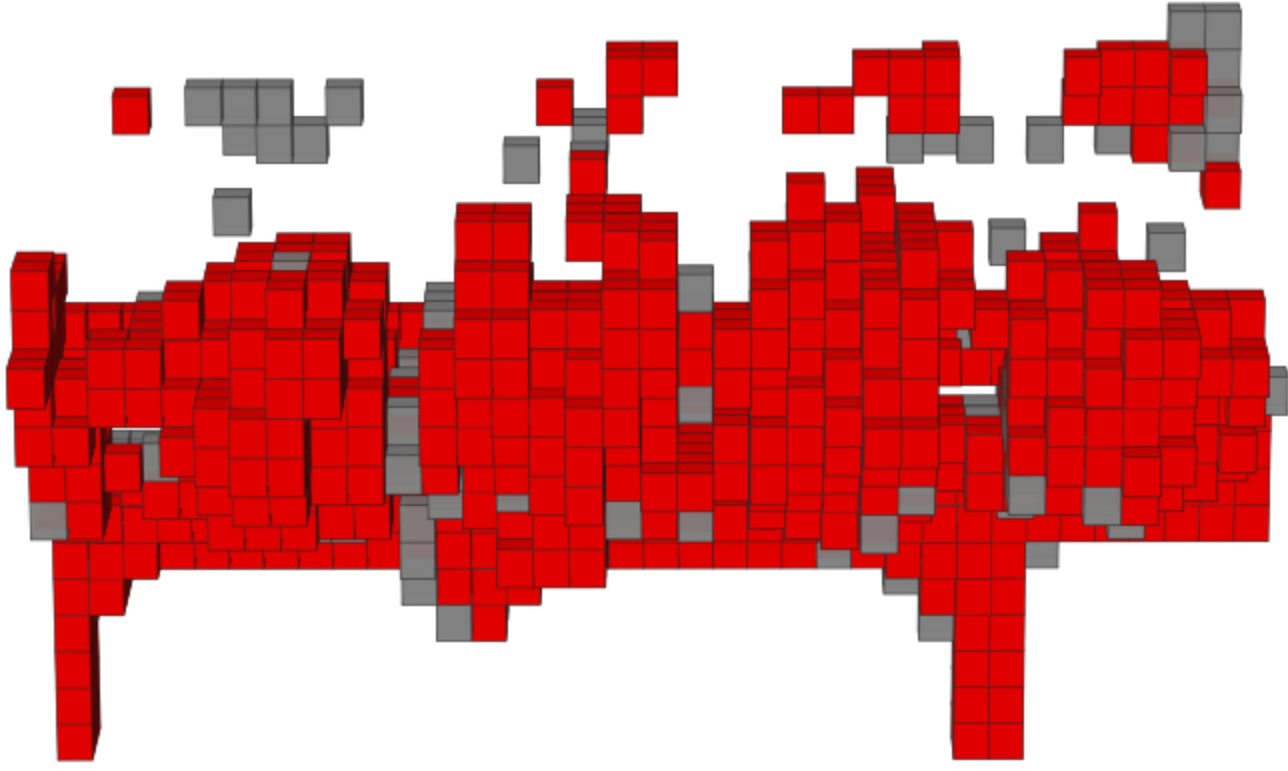}
    \label{fig:g32_d}
  \end{subfigure}

  \begin{subfigure}[t]{0.22\textwidth}
    \centering
    \includegraphics[width=\textwidth]{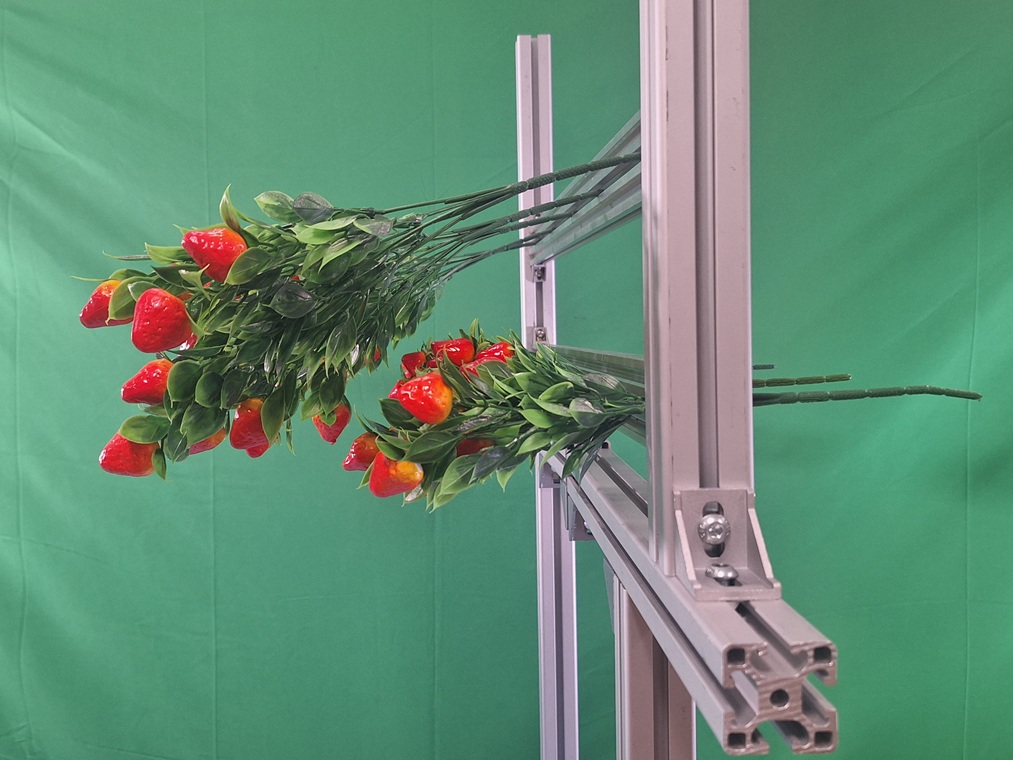}
    \caption{}
    \label{fig:g32_e}
  \end{subfigure}
  \hfill
  \begin{subfigure}[t]{0.22\textwidth}
    \centering
    \begin{tikzpicture}
        \node[inner sep=0] (img) at (0,0)
            {\includegraphics[width=0.6\textwidth]{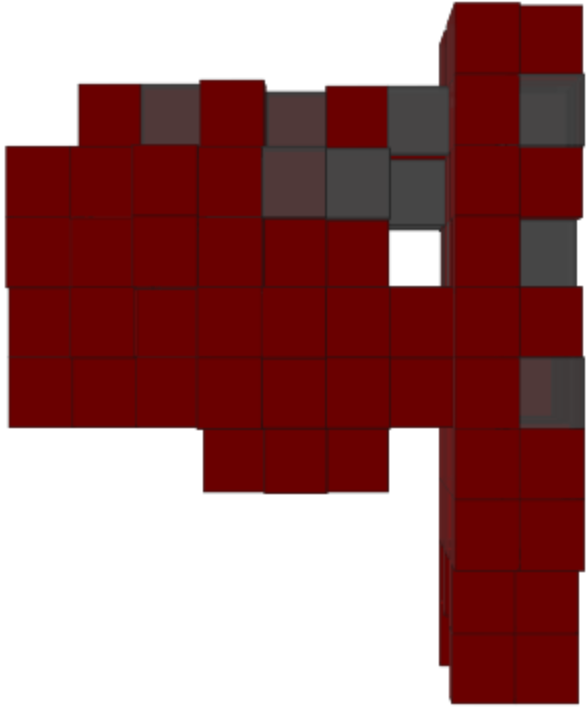}};

        \node[
            anchor=south west,
            fill=white,
            draw=black,
            line width=0.3pt,
            inner sep=1pt,
            font=\scriptsize,
            overlay
        ] at ([xshift=-20pt,yshift=4pt]img.south west) {
            \shortstack[l]{
                $R_\infty= 98.4 \pm 0.1\%$\\[-1pt]
                $T_e = 27.9 \pm 5.4\,\mathrm{s}$
            }
        };
    \end{tikzpicture}
    \caption{}
    \label{fig:g32_f}
  \end{subfigure}
  \hfill
  \begin{subfigure}[t]{0.22\textwidth}
    \centering
    \begin{tikzpicture}
        \node[inner sep=0] (img) at (0,0)
            {\includegraphics[width=0.6\textwidth]{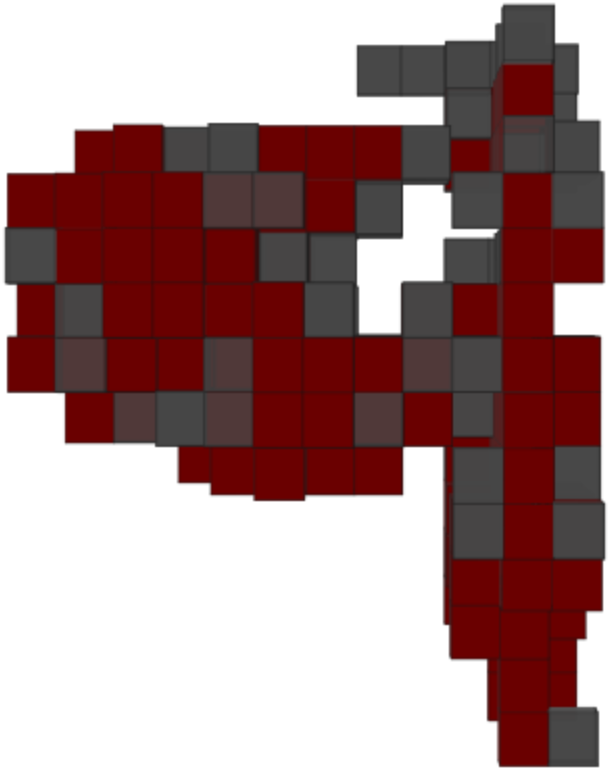}};

        \node[
            anchor=south west,
            fill=white,
            draw=black,
            line width=0.3pt,
            inner sep=1pt,
            font=\scriptsize,
            overlay
        ] at ([xshift=-20pt,yshift=4pt]img.south west) {
            \shortstack[l]{
                $R_\infty = 98.3 \pm 0.3\%$\\[-1pt]
                $T_e = 31.1 \pm 6.0\,\mathrm{s}$
            }
        };
    \end{tikzpicture}
    \caption{}
    \label{fig:g32_g}
  \end{subfigure}
  \hfill
  \begin{subfigure}[t]{0.22\textwidth}
    \centering
    \begin{tikzpicture}
        \node[inner sep=0] (img) at (0,0)
            {\includegraphics[width=0.63\textwidth]{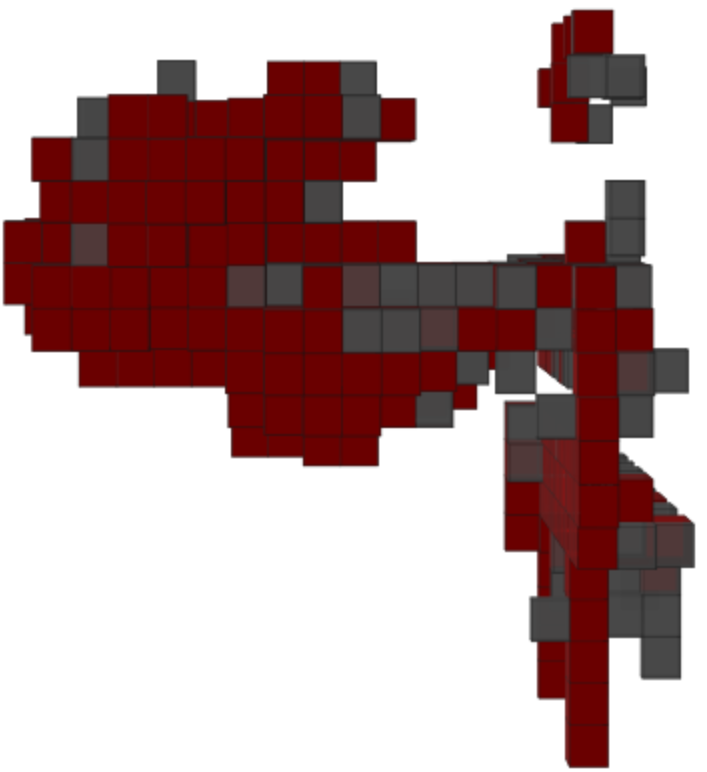}};

        \node[
            anchor=south west,
            fill=white,
            draw=black,
            line width=0.3pt,
            inner sep=1pt,
            font=\scriptsize,
            overlay
        ] at ([xshift=-20pt,yshift=4pt]img.south west) {
            \shortstack[l]{
                $R_\infty = 98.7 \pm 0.4\%$\\[-1pt]
                $T_e = 47.4 \pm 12.0\,\mathrm{s}$
            }
        };
    \end{tikzpicture}
    \caption{}
    \label{fig:g32_h}
  \end{subfigure}

  \caption{Real and mapped examples from the strawberry clusters experiment showing a front view at the top and a side view at the bottom. As the voxel resolution decreases from (b) $4$~cm to (c) $3$~cm to (d) $2$~cm, the robot becomes capable of extracting finer details of the map at the expense of slower exploration.}
  \label{fig:straw_maps}
\end{figure*}

Fig. \ref{fig:shelves_results} provides a direct comparison between the real environment and its voxel-based mapping. Shelves and boxes are correctly placed and sized. Table \ref{tab:exp_shelf_kpi} summarizes the quantitative results of these experiments. 
The robot managed to successfully explore more than $92\%$ of the environment (averaged across 10 experiment) within 120 seconds, with a total movement of less than $14$~m, demonstrating the effectiveness of our algorithm. 

 Explored proportion rate $R(t)$ across all experimental runs is provided in Fig. \ref{fig:shelves_results}b. This shows that the robot regularly scans more than three quarters of the workspace in around 70 seconds. The exploration rate remained consistent across all experimental runs. The robot completed the two minutes exploration without hitting any obstacles despite performing complex maneuvers and regularly reaching into the shelves to scan target voxels. The Supplemental Video demonstrates examples of these complex movements. Obstacles were always avoided even when they were discovered while the robot was moving towards the target voxel. 

Our algorithm required an average of $47$ targets to complete the exploration. Out of these, around $12$ targets were triggered by a discovery rate deadlock, which is illustrated in Fig. \ref{fig:shelves_results}c, while stagnation deadlocks caused around the exploration to change the target around $7$ times on average. These numbers highlight the importance and effectiveness of the deadlock
monitoring module. Without it, the robot may not be able to recover from local minima. Nonetheless, the robot spent about $41\%$ of the exploration time in stagnation, indicating the need for a faster recovery from deadlocks.

\subsection{Simulation tests with cluttered environment}

We tested four scenarios of increasing complexity, defined by obstacles of varying sizes and quantities as shown in Fig. \ref{fig:abc}. The workspace volume was the same of the shelving unit experiment. The goal was to evaluate the ability of the robot to explore cluttered environments. For each scenario, we considered five randomly generated obstacle placements and repeated each placement 10 times. 

Scenario A represents a lightly cluttered environment with two $15$~cm cubes. Scenario B represents moderate clutter with five $12$~cm cubes. Scenario C is a highly cluttered environment with ten $10$~cm cubes. A larger number of smaller cubes increases mutual occlusions and creates a more challenging space to navigate. The robot must maneuver into the cluttered region while actively discovering and avoiding obstacles. These three setups are represented in Fig. \ref{fig:abc}a-c. 

Scenario D tests the challenging case where large foreground objects hide smaller rear objects. Three pairs of cubes are randomly placed in the scene. Each pair consists of a large $20$ cm foreground cube and a smaller $10$ cm rear cube, as shown in Fig. \ref{fig:abc}d. In this scenario, achieving a high exploration coverage $R_\infty$ does not necessarily indicate a successful exploration strategy unless the three occluded rear cubes are discovered and classified as occupied.

\begin{figure}[t!]
  \centering
  \vspace{2.2mm}

  \begin{subfigure}[t]{0.24\textwidth}
    \centering
    \includegraphics[width=0.9\textwidth]{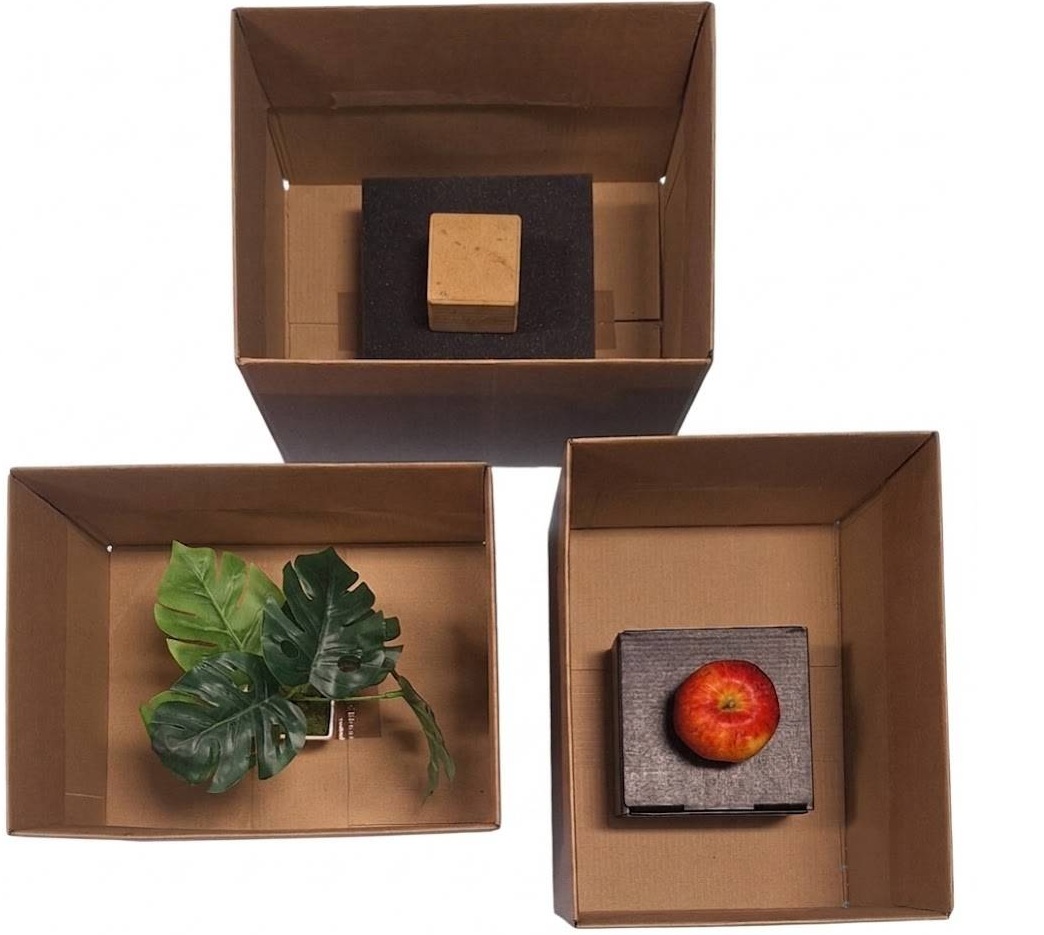}

    \caption{}
  \end{subfigure}%
  \hfill
  \begin{subfigure}[t]{0.24\textwidth}
    \centering
    \begin{tikzpicture}
      \node[inner sep=0] (img) at (0,0)
        {\includegraphics[width=\textwidth]{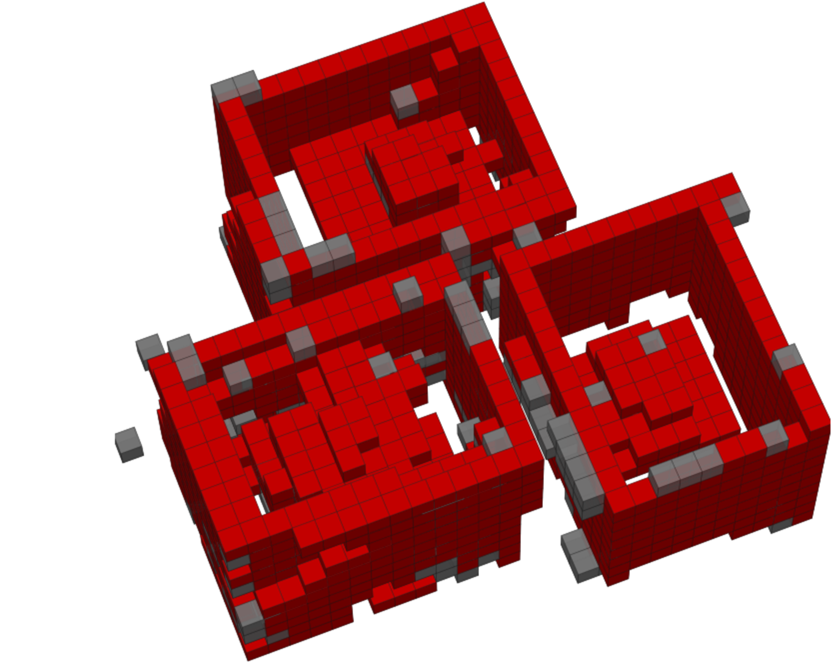}};
    
    \node[
        anchor=south east,
        fill=white,
        draw=black,
        line width=0.3pt,
        inner sep=1pt,
        font=\scriptsize,
        overlay
      ] at ([xshift=27pt,yshift=81pt]img.south west) {
        \shortstack[l]{
          $R_\infty = 96.4 \pm 0.9\%$\\[-1pt]
          $T_e = 85.2 \pm 31.3\,\mathrm{s}$
        }
      };
    \end{tikzpicture}
    \caption{}
  \end{subfigure}

  \caption{Real vs mapped open-top boxes environment from one experimental run.}
  \label{fig:real_boxes_map}
\end{figure}

Fig. \ref{fig:combined_results} shows the final results for these simulations. Across all scenarios, the robot explored most of the workspace, including rear occluded objects in Scenario D, with $R_\infty$ exceeding $94\%$ on average. However, the completion time reveals that the robot needs more time to explore environments with higher clutter density. This is reflected in the percentage of explored workspace shown in Fig. \ref{fig:combined_results}a, which demonstrates slower coverage with increased clutter. The robot scans the majority of the workspace during the first minute and spends around $45\%$ of the exploration time in stagnation, aligning with the results obtained in the shelving unit environment. Fig. \ref{fig:combined_results}c shows a mapped environment from Scenario D, demonstrating the ability to explore occluded regions behind the front cubes.

\section{Exploration of Open Environments}

We conduct two further experiments to assess the performance in 
open environment exploration, looking at the practical cases of exploring open boxes and reconstructing (artificial) strawberry plants. Both are of practical interests and contain occlusions that can only be explored with a robotic arm that has an eye-in-hand camera configuration. 

The voxel reconstruction of strawberry plants is illustrated in Fig. \ref{fig:straw_maps} for different voxel resolutions. As the voxel resolution becomes smaller, the robot manages to extract finer environment details, as clarified by the growing ``gap'' on the side view illustrations (bottom row of Fig. \ref{fig:straw_maps}b-d). Smaller voxel resolution did not cause a significant change in the exploration proportion $R_\infty$ as the robot covered more than $98\%$ of the workspace across all resolutions. However, the exploration was slower, with $T_{e}$ increasing from $27.9$~s for the $4$~cm resolution to $47.4$~s for the $2$~cm one. As shown in the Supplemental Video, high $R_\infty$ is achieved through complex movements to explore the occluded region under the top strawberry clusters.

The voxel reconstruction of the open boxes is illustrated in Fig. \ref{fig:real_boxes_map}. With a voxel resolution of $2.5$~cm,the robot was able to reconstruct the volume of each box content, as shown in Fig. \ref{fig:real_boxes_map}b, with a final exploration coverage exceeding $96\%$. Due to the camera's depth range, these fine details can only be captured at close proximity, a task our algorithm performs safely even inside the box, as demonstrated in the Supplemental Video. 

\section{Discussion \& Comparison}

By combining a virtual model controller with a scoring criterion for targeting voxels in the unknown space, our algorithm drives the robot towards unexplored regions and enables whole-body obstacle avoidance as the robot discovers new obstacles. With these features, our algorithm demonstrated total coverage that exceeded $90\%$ in less than $120$~s across all confined environments and $95\%$ in all open environments, making it suitable for navigating inside tight spaces and exploring hidden and occluded regions.

\begin{figure}[t!]
  \centering
  \vspace{2.2mm}
  \includegraphics[width=0.48\textwidth]{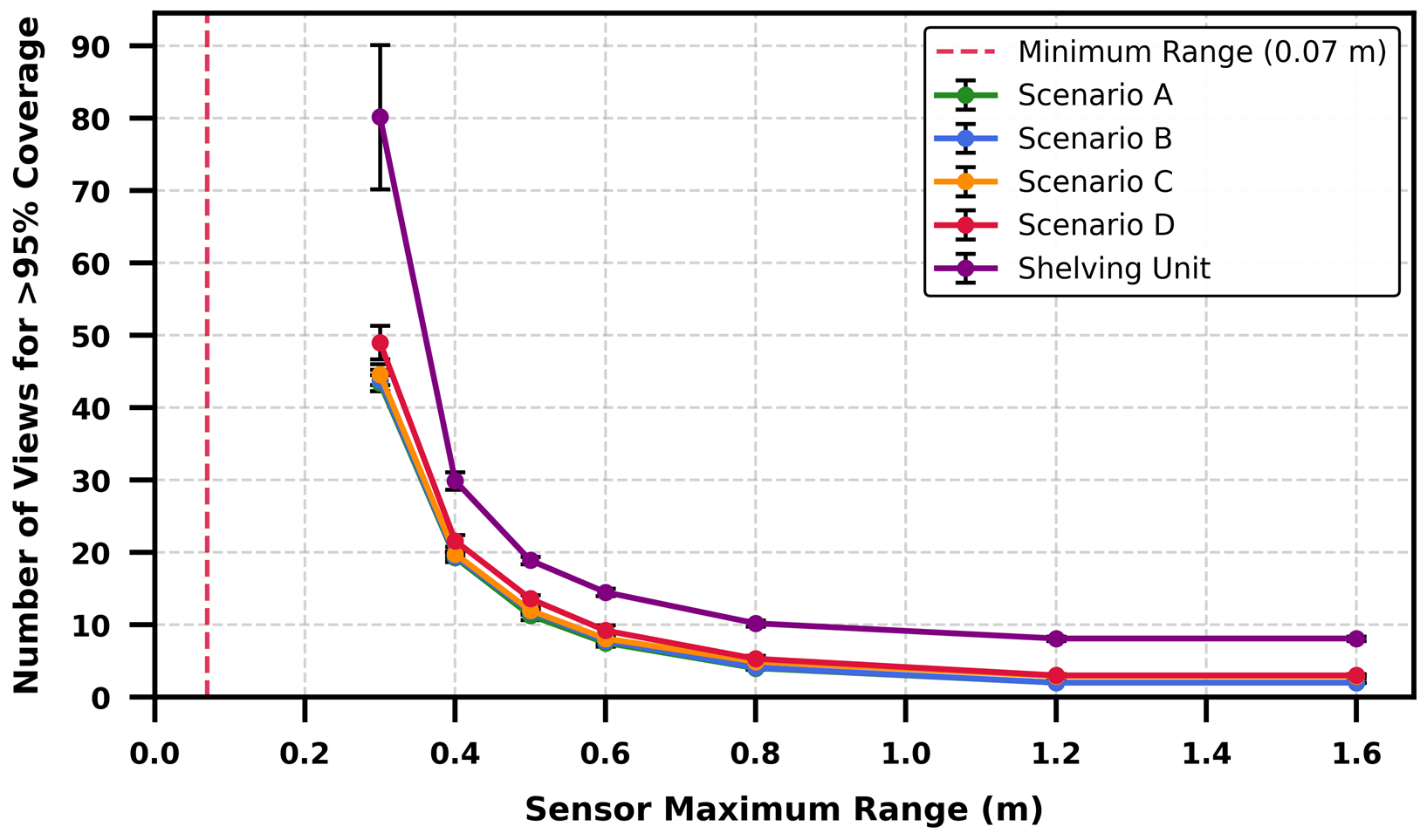}
  \caption{Number of sampled next-best-views required to reach more than $95\%$ of workspace exploration for longer sensor range in Scenarios A, B, C, and D and the shelving unit environments.}
  \label{fig:nbv}
\end{figure}

To highlight the advantages of our algorithm against next-best-view methods, we simulated the standard next-best-view based on information gain for the three-tier shelving unit experiment and the four scenarios of the cluttered environments. The goal is to understand how next-best-view sampling scales with more complex environments and smaller sensor range. We select the views that maximize the information gain while being in the range of motion of the robotic arm. For simplicity, we do not enforce any collision avoidance nor we consider any kinematic feasibility. To focus on range effects, we adopt the minimum range and field of view of the Intel D405 camera ($0.07$~m and $87^\circ \times 58^\circ$) across all simulations. We then vary the maximum range. 
For each range, we repeated the simulation ten times. We halt the simulation if the coverage exceeds $95\%$.

The results demonstrate three critical limitations of next-best-view methods in confined and cluttered environments:
\begin{enumerate}
    \item \textit{View count scaling:} With more clutter and reduced sensor range, the number of views required to cover more than $95\%$ of the workspace significantly increases, as seen in Fig. \ref{fig:nbv}, which is a direct result of the lower observable volume per view and higher occlusion. In contrast, our proposed algorithm abandons the next-best-view concept in favor of directly selecting target voxels in the unknown space, allowing information to be accrued continuously along the robot trajectory without explicitly generating and planning to a sequence of views.
    \item \textit{Motion planning feasibility:} A view is not useful unless the robot can reach it. In more cluttered environments, the feasibility of reaching target views plummets \cite{orthey2023sampling}. For example, Li et al. \cite{li2025compass} demonstrated that successful motion planning toward target views is achieved for $60\%$--$72\%$ out of all generated next-best-views with an optimized planner. This is exacerbated by the high number of views required in confined spaces. Our algorithm mitigates this issue by employing Virtual Model Control for reactive avoidance of obstacles as they are discovered, which eliminates the computational burden of inverse kinematics and global trajectory planning toward next-best-views.
    \item \textit{Planning latency:} Motion planning to each view incurs substantial latency and computational burden. Liu et al. \cite{liu2024manipulability} reported $5$~s of path planning time for each next-best-view, while Ren et al.~\cite{ren2023robot} required more than two minutes of planning for an average of just four views. With increased number of required views to explore confined environments, these methods can become prohibitive for real-time exploration. Compared to that, our algorithm does not require any motion planning and directly moves the robot towards the unknown space by computing the torques arising from the interaction forces with the target and obstacles.
\end{enumerate}

Despite these advantages of our algorithm, several limitations are notable in its current form that are worth exploring in future works. Our scoring method selects the highest-score voxel from all voxels in the scene, which is suitable for small workspaces, such as the environments considered in our experiments, but may become computationally expensive for larger workspaces. Integrating more efficient hierarchical representations with adaptive spatial resolution (e.g., OctoMap \cite{Hornung2013}) into our algorithm may facilitate the exploration of large workspaces. Furthermore, parameter selection is not trivial and has to be carefully considered as certain choices may lead to collisions. Additionally, the algorithm does not consider the joints efforts and manipulability, which can produce undesirable movements. Finally, improving the deadlock
monitoring module may enable earlier detection of deadlocks and expedite the exploration process.

\section{Conclusion}

In this work, we proposed an active environment exploration algorithm designed for mapping confined spaces by enabling redundant robotic arms with eye-in-hand camera configuration to navigate narrow and occluded spaces. The algorithm integrates a scoring method for choosing target voxels directly from the unexplored space with a virtual model controller that drives the robot safely into unknown regions and avoids newly discovered obstacles without the need to re-plan paths. Extensive evaluations across a three-tier shelving unit, simulated cluttered scenarios, open filled boxes, and multi-layer strawberry plantation demonstrated that our method reliably maps over $90\%$ of workspaces in under $120$~s while avoiding collisions. Future work will focus on improving the deadlock mitigation approach to reduce stagnation time and integrating the algorithm as a local planner within a global next-best-view method.

\bibliographystyle{IEEEtran}
\bibliography{references}

\end{document}